\documentclass[10pt,twocolumn,letterpaper]{article}

\usepackage{cvpr}              

\usepackage[utf8]{inputenc} 
\usepackage[T1]{fontenc}    
\usepackage{url}            
\usepackage{booktabs}       
\usepackage{amsfonts}       
\usepackage{nicefrac}       
\usepackage{microtype}      
\usepackage{amsmath}
\usepackage{mathtools}
\usepackage{physics}
\usepackage{bm}
\usepackage{glossaries}
\usepackage{duckuments}
\usepackage{adjustbox}
\usepackage[normalem]{ulem}
\usepackage[dvipsnames]{xcolor}

\newcommand{\TODO}[1]{\textbf{\color{red}[TODO: #1]}}

\newcommand{\modv}{{\color{BurntOrange}V}}
\newcommand{\modav}{{\color{OliveGreen}AV}}
\newcommand{\vul}[1]{{\color{BurntOrange}\uline{\color{black}#1}}}
\newcommand{\avul}[1]{{\color{OliveGreen}\uline{\color{black}#1}}}

\renewcommand{\TODO}[1]{}

\newacronym{cnn}{CNN}{Convolutional Neural Network}
\newacronym{nlp}{NLP}{Natural Language Processing}
\newacronym{vit}{ViT}{Vision Transformer}
\newacronym{mae}{MAE}{Masked Autoencoder}

\definecolor{cvprblue}{rgb}{0.21,0.49,0.74}
\usepackage[pagebackref,breaklinks,colorlinks,citecolor=cvprblue]{hyperref}
\usepackage{multirow,multicol}

\def\paperID{10840} 
\def\confName{CVPR}
\def\confYear{2024}

\title{AVFF: Audio-Visual Feature Fusion for Video Deepfake Detection}

\newcommand{\authorskip}{\hspace{8mm}}

\author{Trevine Oorloff\textsuperscript{1,2\thanks{This work was completed during an internship at Reality Defender Inc.}}
\authorskip Surya Koppisetti\textsuperscript{2}
\authorskip Nicol\`o Bonettini\textsuperscript{2}
\authorskip Divyaraj Solanki\textsuperscript{2} \\ 
\authorskip Ben Colman\textsuperscript{2}
\authorskip Yaser Yacoob\textsuperscript{1}
\authorskip Ali Shahriyari\textsuperscript{2}
\authorskip Gaurav Bharaj\textsuperscript{2}
\vspace{3mm}\\ 
\textsuperscript{1}University of Maryland - College Park \hspace{8mm} \textsuperscript{2}Reality Defender Inc.\\
}

\begin{document}
\maketitle
\begin{abstract}
With the rapid growth in deepfake video content, we require improved and generalizable methods to detect them. 
Most existing detection methods either use uni-modal cues or rely on supervised training to capture the dissonance between the audio and visual modalities. 
While the former disregards the audio-visual correspondences entirely, the latter predominantly focuses on discerning audio-visual cues {\em within the training corpus}, thereby potentially overlooking correspondences that can help detect unseen deepfakes.
We present Audio-Visual Feature Fusion (AVFF), a two-stage cross-modal learning method that explicitly captures the correspondence between the audio and visual modalities for improved deepfake detection. The first stage pursues representation learning via self-supervision on real videos to capture the intrinsic audio-visual correspondences. To extract rich cross-modal representations, we use contrastive learning and autoencoding objectives, and introduce a novel audio-visual complementary masking and feature fusion strategy. The learned representations are tuned in the second stage, where deepfake classification is pursued via supervised learning on both real and fake videos.
Extensive experiments and analysis suggest that our novel representation learning paradigm is highly discriminative in nature. 
We report 98.6\% accuracy and 99.1\% AUC on the FakeAVCeleb dataset, outperforming the current audio-visual state-of-the-art by 14.9\% and 9.9\%, respectively. 
\end{abstract}    
\vspace{-1em}
\section{Introduction}
\label{sec:intro}

Deepfake generative AI technology enables new opportunities to create rich and quality content in multimedia applications such as virtual reality~\cite{wu2023deepfake}, movie production~\cite{prajwal2020wav2lip}, and telepresence~\cite{Ma_2021_CVPR, wang2021one}. However, its malicious use has become a major societal threat posing a number of problems including frauds\footnote{\href{https://www.wsj.com/articles/fraudsters-use-ai-to-mimic-ceos-voice-in-unusual-cybercrime-case-11567157402?_sm_au_=i5VjqL43tSrQV756QcLJjKQ1j7GJ1}{Fraudsters Used AI to Mimic CEO’s Voice in Cybercrime Case}}, defamation\footnote{\href{https://www.bbc.com/news/entertainment-arts-65854112}{Deepfake porn documentary explores its `life-shattering' impact}, \href{https://www.washingtonpost.com/technology/2023/11/05/ai-deepfake-porn-teens-women-impact/}{AI Fake Nudes are booming}} and disinformation\footnote{\href{https://www.bbc.com/news/uk-66130785}{Martin Lewis felt `sick' seeing deepfake scam ad on Facebook}, \href{https://www.nbcnews.com/tech/tech-news/deepfake-scams-arrived-fake-videos-spread-facebook-tiktok-youtube-rcna101415}{Deepfake scams have arrived}}. As the generative AI landscape continues to evolve, there is a growing need for robust deepfake detection that helps preserve content integrity.
In this paper, we study video deepfake detection where either or both the visual and audio content are AI-generated. 

\begin{figure}[t]
    \centering
    \includegraphics[width=0.86\linewidth]{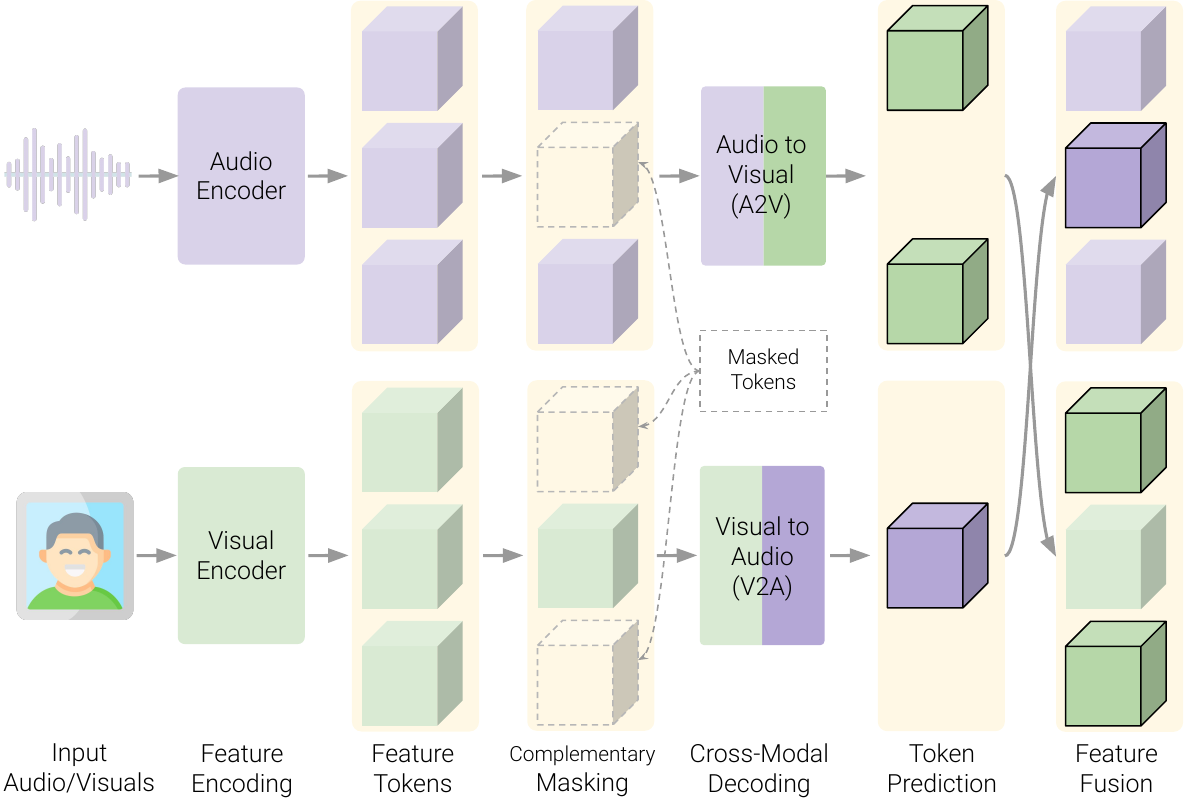}
    \caption{We use audio-visual correspondences for deepfake detection. Transformer-based encoders are used to extract audio and visual feature tokens, which are then masked complementarily. The visible audio tokens are sent through a learnable A2V network to predict the masked visual tokens. {These predicted visual tokens are fused with the visible visual tokens to obtain the full visual embeddings. Full audio embeddings are obtained in a similar way using the V2A network.} 
    The audio/visual embeddings are then used for video reconstruction in the MAE sense, and subsequently for deepfake classification.}
    \label{fig:teaser}
    \vspace{-1em}
\end{figure}

We pursue multi-modal learning and draw inspiration from previous works, such as SyncNet~\cite{Chung16a-syncnet}, CLIP~\cite{radford2021clip}, and AudioCLIP~\cite{guzhov2022audioclip}, where the correspondence between different modalities (audio, text, visual) was leveraged to significantly enhance performance on various downstream tasks. We note that in real video face context, the audio-visual correspondence is deeply intuitive since there is an intrinsic correlation between the mouth articulations (visemes) and the speech units (phonemes)~\cite{zhou2018visemenet,agarwal2020detecting,prajwal2020wav2lip,Chung16a-syncnet}, as well as an alignment of emotional nuances embedded in the facial and speech expressions~\cite{mittal2020emotions,mittal2020m3er, bai2023aunet}. Such inherent audio-visual correspondence, for example, in audio-driven emotion,  is challenging to faithfully replicate in deepfake videos. Based on these observations, we propose a video deepfake detection method that learns efficient representations for audio and visual modalities. The proposed method employs a novel complementary masking and cross-modal feature fusion strategy to explicitly capture the audio-visual correspondences.

Previous literature on audio-visual video deepfake detection \cite{yang2023avoiddf, mittal2020emotions} use supervised contrastive learning to capture the audio-visual correspondence. Such methods align the audio and visual embeddings closer to each other, if the content in both modalities is real, and push them apart if either or both modalities are generative. Similarly, others pursue a single stage supervised learning method, where models are directly trained on labeled deepfake datasets for deepfake classification~\cite{khalid2021fakeavceleb, dolhansky2020dfdc}.  
While such methods yield promising results, we conjecture that they may not fully exploit the audio-visual correspondence.  Also, training solely on a deepfake dataset narrows the model's focus to discern separable features within the training corpus, potentially overlooking subtle audio-visual correspondences that can help detect unseen deepfake samples (observe in \cref{tab:intra_eval,tab:cross_data_manipul}, the weaker performance of other baselines compared to \cite{haliassos2022lrealforensics} and ours). 

To circumvent these issues, we propose a two stage training pipeline comprising of (i) a self-supervised representation learning stage that explicitly enforces audio-visual correspondence using a novel approach, and (ii) a supervised downstream classification stage. 
In the representation learning stage, we extract audio-visual representations via self-supervised learning on real face videos, which are available in abundance \cite{wolf2011ytfaces, zhu2022celebvhq, afouras2018lrs3}. Drawing inspirations from CAV-MAE \cite{gong2023cavmae}, we make use of the complementary nature of two learning objectives: contrastive learning and autoencoding. For extracting rich representations, we supplement the contrastive learning objective by a novel audio-visual complementary masking and fusion strategy that sits within the autoencoding objective. In the classification stage, we train a classifier that exploits the lack of cohesion between audio-visual features of deepfake videos to separate them from real videos.

We evaluate our method against existing state-of-the-art approaches on multiple benchmarks. Our results reveal substantial improvements, when compared against the existing audio-visual state-of-the-art, enhancing the performance by 9.9\% in AUC and 14.9\% in accuracy when evaluated on the FakeAVCeleb dataset~\cite{khalid2021fakeavceleb}.
This underscores the effectiveness of {\em explicitly} leveraging audio-visual correspondences through the proposed method. In summary:
\begin{itemize}
\item We propose a novel self-supervised representation learning method that {\em explicitly} captures audio-visual correspondences in real videos. To learn the correspondences, we pursue a dual-objective of contrastive learning and autoencoding, and supplement it with a novel audio-visual complementary masking and fusion strategy.

\item Qualitative analysis using t-SNE~\cite{Maaten2008VisualizingDU} shows a clear separation between the real and fake video embeddings at the end of the representation learning stage. This demonstrates the efficacy of the proposed representation learning. 

\item We propose a two-stage deepfake detection method comprising of the aforementioned representation learning stage followed by a deepfake classification stage. Our method yields state-of-the-art performance on deepfake detection when either or both the audio and visual contents are AI generated. 
We achieve 98.6\% accuracy and 99.1\% AUC on FakeAVCeleb, surpassing the existing audio-visual state-of-the-art by 14.9\% and 9.9\% respectively.

\end{itemize}

\section{Related Works}
\label{sec:rw}

\subsection{Multi-Modal Representation Learning} 

Learning a joint representation from multiple modalities has been shown to be effective for different tasks in the state-of-the-art. SyncNet~\cite{Chung16a-syncnet} proposes a Siamese Network to estimate the lip-sync error between audio and visual modalities. This framework processes each modality through a distinct branch and employs a contrastive loss to promote the similarities in the encoding space. More recently, improvements in the \gls{nlp} field brought by BERT~\cite{Devlin2019BERTPO}, allowed to use text modality in multi-modal frameworks. Another example is CLIP~\cite{radford2021clip}, a zero-shot image classification model that leverages separate encoders for images and captions to find a suitable pairing in the latent space. AudioCLIP~\cite{guzhov2022audioclip} extends this approach to audio, enabling multi-modal classification.

Several self-supervised methods have emerged, inspired by the \gls{mae} framework~\cite{kaiming2022mae}. 
AV-MAE \cite{georgescu2023avmae} is a joint masked autoencoder for audio, visual, and joint audio/visual classification. 
The authors explore different encoding policies for dual-modality inputs, demonstrating the ability to decode one masked modality from the other. 
CAV-MAE~\cite{gong2023cavmae} raises concerns about the ability of a vanilla masked autoencoder to learn a coordinated representation between audio and visuals (\ie, a representation that enforces similarity~\cite{baltrusaitis2019multimodal}) and adds a contrastive loss to explicitly leverage the audio-visual pair information. 
In this work, we draw inspiration from CAV-MAE, in using a dual contrastive-autoencoding objective for effective representation learning. Our approach diverges from existing MAE literature: (i) in terms of the masking strategy, where we use a complementary masking strategy post-encoding; (ii) in terms of the cross-modal fusion, where for every modality we replace the shared learnable masked tokens of MAEs with tokens predicted from the other modality. We do this to enforce explicit correspondence between audio and visual modalities. 
\looseness=-1

\begin{figure*}
    \centering
    \includegraphics[width=0.925\linewidth]{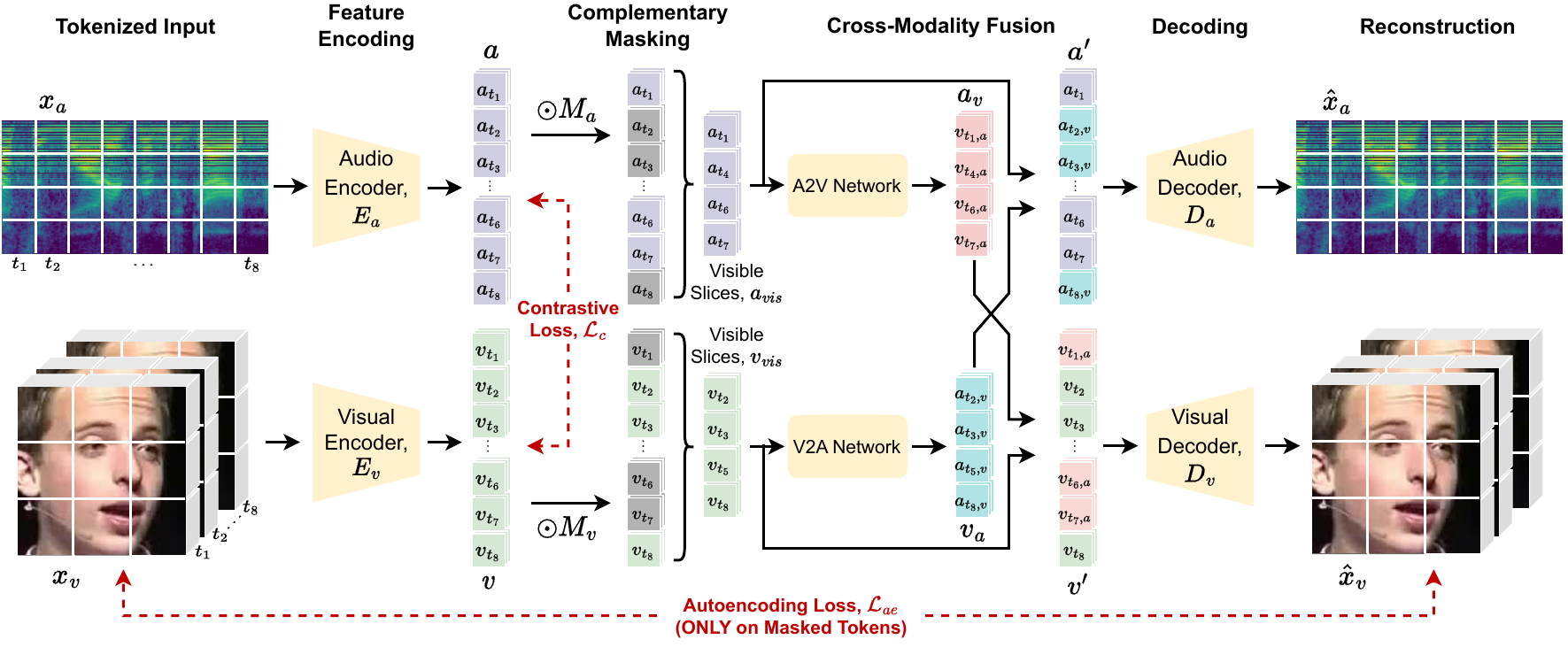}
    \vspace{-0.5em}
    \caption{\textbf{Audio-Visual Representation Learning Stage.} A real input sample, $x \in \mathcal{D}_r$, with corresponding audio and visual tokens ($\bm{x_a}$, $\bm{x_v}$), is split along the temporal dimension, creating $K$ slices, 
    $\{x_{a,t_i}\}_{i=1}^K$ and $\{x_{v,t_i}\}_{i=1}^K$ (illustrated with $K=8$ in the figure). 
    The temporal slices are then encoded using unimodal transformers, $E_a$ and $E_v$, to yield feature embeddings $\bm{a}$ and $\bm{v}$. We then complementarily mask $50\%$ of the temporal slices in ($\bm{a}$, $\bm{v}$)  with binary masks ($\bm{M}_a$, $\bm{M}_v$).
    The visible slices of $\bm{a}$ and $\bm{v}$ are passed through A2V and V2A networks respectively, to generate cross-modal slices $\bm{a_v}$ and $\bm{v_a}$. The masked slices of $\bm{a}$ and $\bm{v}$ are then replaced with the corresponding slices in $\bm{a_v}$ and $\bm{v_a}$. The resulting cross-modal fusion representations, $\bm{a'}$ and $\bm{v'}$, are input to unimodal decoders to obtain the audio and visual reconstructions, $\bm{\hat{x}_a}$ and $\bm{\hat{x}_v}$. For the learning, we use a dual-objective loss function, which computes the contrastive loss between the audio and visual feature embeddings and the autoencoder loss between the input and the reconstruction of the masked tokens. 
    }
    \label{fig:pipeline_rl}
    \vspace{-1em}
\end{figure*}

\subsection{Deepfake Detection} 

\noindent\textbf{Visual-only methods.}
Multiple recent works have made use of visual-only artifacts for deepfake detection.
The authors of ~\cite{rossler2019faceforensics++} train a \gls{cnn} (XceptionNet) end-to-end, setting one of the first baselines on the dataset they shared with the community.  
Some methods target specific face regions for exposing deepfakes. For example, LipForensics~\cite{haliassos2021lipsdontlie} relies on lip movements that might be difficult to reproduce by generative methods. Others consider inconsistent head pose~\cite{yang2019head, lutz2021head} or eye blinking~\cite{Li2018eye, Jung2020eye}.
Another common approach is to consider both spatial and temporal domains. FTCN~\cite{zheng2021ftcn} proposes a combination of a \gls{cnn} and a transformer network to exploit short-time and long-time temporal incoherence. Similarly, \cite{zhao2022spatiotemporal} extracts spatial features by means of an attention-based network and then fuses those features with a temporal module.

Several papers based on the \gls{vit} have been published since the advent of the original paper~\cite{dosovitskiy2020vit} for image classification. An example is CViT~\cite{wodajo2021cvit}, where learnable features are extracted by means of a \gls{cnn} and subsequently fed to a \gls{vit} for the classification task. Similar approaches are followed by~\cite{kaddar2021hcit, zhuang2022uiavit, Dong2022protecting}.
Recently, generalization to unseen deepfake methods~\cite{Ojha_2023_CVPR, Wang_2023_CVPR} and the impact of the identity leakage during training~\cite{Dong_2023_CVPR, Huang_2023_CVPR} have also been investigated. 
RealForensics~\cite{haliassos2022lrealforensics} proposed a hybrid approach that consists of using a multi-modal pre-training pipeline, where audio and visuals exclusively from real samples are used for computing internal representations that help the classifier to discriminate between real and fake video. This is not considered a pure multi-modal approach as the final classification is performed just on visuals, discarding the audio modality. 
As modern-day video deepfakes consist of both audio and visual manipulations, uni-modal deepfake detection methods prove to be less effective.

\vspace{.4em}\noindent\textbf{Audio-visual methods.}
These methods consider audio and visual signals to target deepfake detection on both modalities.
One of the first papers to address multi-modality is \textit{Emotions Don't Lie}~\cite{mittal2020emotions}, proposing a Siamese Network where uni-modal features are passed to an emotion recognition network to compare affective cues corresponding to perceived emotion from the two modalities within a video.
\textit{Not made for each other}~\cite{chugh2020mds} explicitly modeled the dissimilarity between modalities, and proposed the Modality Dissonance Score (MDS) network, where a contrastive loss is computed on single modality embeddings to expose differences on audio-visual pairs. 
\textit{Voice-Face matching Detection} (VFD)~\cite{cheng2022vfd} is another example of using contrastive loss for modeling face and voice homogeneity.
A similar concept is exploited in \cite{agarwal2020detecting}, where the focus is on phoneme-viseme mismatch. The idea is that a given dynamic of mouth shape (viseme) should correspond to a given emitted sound (phoneme). The authors only focus on the mouth region, showing how deepfake methods struggle to reproduce certain dynamics.

More recently, the paradigm for multi-modality shifted towards fusion of single-modality features. AV-DFD~\cite{zhou2021avdfd} proposes a joint audio-visual deepfake detection framework in which visual and audio features are aligned and tiled to be passed onto a cross-attention mechanism on the temporal dimension. More recent papers study the encoding/decoding potential of \gls{vit}s and build feature fusion in the embedding space on the decoder side. Examples are AVFakeNet~\cite{ilyas2023avfakenet} and AVoiD-DF \cite{yang2023avoiddf}.

\section{Method}\label{sec:method}

The proposed algorithm, AVFF, consists of two stages: (i) representation learning, and (ii) deepfake classification. Stage 1 aims to acquire an audio-visual representation with cross-modal correspondence via self-supervised learning, and it solely utilizes real face videos. The model learns audio-visual correspondences inherent to real videos via a contrastive learning objective and an effective complementary masking and fusion strategy that sits within an auto-encoding objective (see Fig. \ref{fig:pipeline_rl}). Here, the complementary masking and fusion strategy takes uni-modal audio and visual embeddings $(\bm{a}, \bm{v})$ and systematically masks them to force the learning of advanced embeddings $(\bm{a'}, \bm{v'})$ via reconstruction in the MAE sense. To instill cross-modal dependency, tokens from one modality are used to learn the masked embeddings of the other modality via cross-modal token conversion networks (A2V and V2A in Fig. \ref{fig:pipeline_rl}). Since we work exclusively with real face videos in this stage, the model learns the dependency between ``real'' speech audio and the corresponding visual facial features. In Stage 2, a classifier is trained to distinguish between real and fake videos using the learned representations from the first stage. Effectively, the representation learning stage serves as pre-training for the downstream task of video deepfake detection.

\subsection{Preprocessing} \label{subsec:preprocessing}
Initially, visual frames and the corresponding audio waveforms are extracted from raw videos at a sampling rate of 5~fps and 16~kHz, respectively. Given our emphasis on audio-visual correspondence, we align the cropped facial regions and eliminate the background in the visual frames using  FaceX-Zoo~\cite{wang2021facex}.
This step is performed since background variations typically exhibit minimal correspondence with speech audio. Simultaneously, the audio waveform is converted into a log-mel spectrogram with $L$ frequency bins.
Henceforth, we refer to the preprocessed visual frames and log-mel spectrograms as the visual and audio input representations, respectively.

\subsection{Representation Learning Stage}\label{subsec:rep_learn}
The primary objective of this stage is to learn a representation that effectively captures audio-visual feature correspondences inherently present in real videos (and different thereof from the correspondences in fake videos). Drawing inspiration from CAV-MAE~\cite{gong2023cavmae}, we propose a dual self-supervised learning approach that incorporates contrastive learning and autoencoding objectives. While contrastive learning helps build cross-modal correlations~\cite{radford2021clip, guzhov2022audioclip, morgado2021robust, rouditchenko2021avlnet}, we found in preliminary experiments that relying solely on it does not establish a strong correspondence between the audio and visual modalities. Therefore, we supplement it with an autoencoding objective and embed a complementary masking and cross-modal fusion strategy into the autoencoding framework. This allows us to learn rich cross-modal representations that result in improved deepfake detection (see \cref{tab:ablations}). In \cref{fig:pipeline_rl}, we illustrate the overall pipeline for our representation learning stage and discuss its key components next.\looseness=-1

\vspace{.4em}\noindent\textbf{Input Tokenization.}
Given a dataset $\mathcal{D}_r$ of real talking human portrait videos, we denote a video sample $x \in \mathcal{D}_r$ with a time duration $T$ as comprising of audio and visual components, ${x}_a \in \mathbb{R}^{T_a \times L}$ and  ${x}_v \in \mathbb{R}^{T_v \times C \times H \times W}$, respectively. In ${x}_a$, the $(T_a, L)$ denote the number of audio frames and mel-frequency bins. In ${x}_v$, the ($T_v$, $H$, $W$, $C$) denote the number of visual frames, height, width, and number of channels. We choose $T_a$ and $T_v$ such that $T_a \cdot n_a = T_v \cdot n_v = T$, where $n_a$ and $n_v$ are the sampling frequencies of the audio and visual sequences. We tokenize ${x}_a$ using  $16\times 16$ non-overlapping 2D patches (similar to Audio-MAE~\cite{huang2022audiomae}), and ${x}_v$ using $2 \times 16 \times 16$ non-overlapping 3D spatio-temporal patches (similar to MARLIN~\cite{cai2023marlin}). The resulting representations are denoted as $\bm{x_a}$ and $\bm{x_v}$. Subsequently, we segment each of the tokenized representations into 8 equal temporal slices, $\bm{x_a} = \{x_{a,{t_i}}\}_{i=1}^8$ and $\bm{x_v} = \{x_{v,{t_i}}\}_{i=1}^8$, where the number of slices was decided empirically. This slicing preserves the temporal correspondence of each slice between modalities, as $x_{a,{t_i}}$ and $x_{v,{t_i}}$ correspond to the same time interval.\looseness=-1 

\vspace{.4em}\noindent\textbf{Feature Encoding.}
The two uni-modal audio and visual encoders, $E_a$ and $E_v$, encode the tokenized inputs, $\bm{x_a}$ and $\bm{x_v}$, and output uni-modal features $\bm{a}$ and $\bm{v}$ respectively: $\bm{p} = \{p_{t_i}\}_{i=1}^8 =  E_p(\bm{x_p} + pos_p^e)$, where, $p\in\{a,v\}$ and $pos_p^e$ is the learnable positional embedding.

\vspace{.4em}\noindent\textbf{Complementary Masking.}
Within the uni-modal feature embeddings, $\bm{a}$ and $\bm{v}$, we mask $50\%$ of the temporal slices using binary masks, $(\bm{M}_a, \bm{M}_v) \in \{\mathbf{0}, \mathbf{1} \}$, such that $\bm{M}_a$ and $\bm{M}_v$ are complementary to each other, \ie, $\bm{M}_a = \mathbf{1}$ for slices where $\bm{M}_v = \mathbf{0}$ and vice-versa. 
In other words, for every masked slice in the audio feature, the corresponding visual slice is visible and vice versa. Let us denote the visible temporal slices as $\bm{p_{vis}} = \bm{M}_p \odot \bm{p}$ and the masked temporal slices as $\bm{p_{msk}} = (\neg \bm{M}_p) \odot \bm{p}$, where $p \in \{a,v\}$, $\odot$ denotes the Hadamard product and $\neg$ is the NOT operator.

\vspace{.4em}\noindent\textbf{Cross-Modal Fusion.}
Next, the {\em visible} temporal slices $\bm{a_{vis}}$ and $\bm{v_{vis}}$ are sent through learnable audio-to-visual (A2V) and visual-to-audio (V2A) networks to create their cross-modal temporal counterparts, $\bm{{v}_a} = \mathrm{A2V}(\bm{a_{vis}})$ and $\bm{{a}_v} = \mathrm{V2A}(\bm{v_{vis}})$, respectively. Here, $\bm{{v}_a}$ contains $ \{{v}_{t_i,a} = \mathrm{A2V}(a_{t_i}), \; \; \forall t_i \; \text{where} \; a_{t_i} \in \bm{a_{vis}}\}$, and similarly $\bm{{a}_v}$. Each of the A2V/V2A networks is composed of a single-layer MLP to match the number of tokens of the other modality followed by a single transformer block. 
The audio embedding $\bm{a}'$ is then created using cross-modal fusion, wherein, we take the original feature $\bm{a}$ and replace each masked slice with the corresponding slice of the same temporal index in the cross-modal vector $\bm{{a}_v}$ given by the V2A network (see \cref{fig:pipeline_rl}). The visual embedding $\bm{v}'$ is similarly obtained from the original feature $\bm{v}$ and the cross-modal feature $\bm{v}_a$.
Effectively, this process replaces the masked temporal slices of each modality with cross-modal slices generated from the corresponding temporal slices in the other modality. This is feasible due to our complementary masking strategy, as the masked slices of one modality are the visible slices of the other modality.\looseness=-1

\vspace{.4em}\noindent\textbf{Decoding.}
The uni-modal audio and visual decoders, $G_a$ and $G_v$, take $\bm{a'}$ and $\bm{v'}$ as input to generate the audio and visual reconstructions, $\bm{\hat{x}_a} = G_a(\bm{a'} + pos^g_a)$ and $\bm{\hat{x}_v} = G_v(\bm{v'} + pos^g_v)$, where $pos^g_a$ and $pos^g_v$ are learnable positional embeddings for each modality. The decoders use a transformer-based architecture and are entrusted with the task of effectively utilizing the mix of uni-modal slices and cross-modal slices present in $\bm{a'}$ and $\bm{v'}$ to generate the reconstructions for the two modalities. 

\vspace{.4em}\noindent\textbf{Loss Functions.}
We use a dual objective loss, comprising an audio-visual contrastive loss and an autoencoding loss. The bi-directional audio-visual contrastive loss is defined as:
\begin{equation}
\footnotesize
    \mathcal{L}_c =   - \mkern-24mu \sum_{ \substack{p,q \in \{a,v\}, \\ p \neq q}} \mkern-12mu \frac{1}{2N} \sum_{i=1}^N\log \left [ \frac{\exp(\|\bm{\bar{p}}^{(i)}\|^T \|\bm{\bar{q}}^{(i)}\|/\tau)}{\sum_{j=1}^N \exp(\|\bm{\bar{p}}^{(i)}\|^T \|\bm{\bar{q}}^{(j)}\|/\tau)}\right ] 
\end{equation}
where 
$\bm{\bar{p}}^{(i)}$
is the mean latent vector across the patch dimension of the uni-modal embeddings of the $i$-th data sample, $N$ is the number of samples, $\tau$ is the temperature parameter, and $i,j$ are sample indices. 
The audio-visual contrastive loss enforces similarity constraints between the audio and visual embeddings of a given sample.   

The autoencoder loss, $\mathcal{L}_{ae}$, is composed of reconstruction and adversarial losses, similar to MARLIN \cite{cai2023marlin}. The reconstruction MSE loss, $\mathcal{L}_{rec}$, is computed between the inputs $(\bm{x_a}$, $\bm{x_v})$, and their reconstructions $(\bm{\hat{x}_a}$, $\bm{\hat{x}_v})$, and is computed only over the masked tokens following the approach in MAEs~\cite{gong2023cavmae, cai2023marlin, huang2022audiomae}:
\begin{equation}
\footnotesize
    \mathcal{L}_{rec} = \sum_{p \in \{a,v\}} \frac{1}{N} \sum_{i=1}^N \| {\bm{x_{p_{msk}}}^{(i)}} - \bm{\hat{x}_{p_{msk}}}^{(i)} \|
\end{equation}
For the adversarial loss, $\mathcal{L}_{adv}$, we use the Wasserstein GAN loss \cite{arjovsky2017wasserstein} to supplement the reconstruction loss by enhancing the features captured in the reconstructions of each modality. Similar to the reconstruction loss, the adversarial loss is computed only on the masked tokens: 
\begin{equation}
\footnotesize
    \mathcal{L}_{adv}^{(G)} = - \mkern-12mu \sum_{p \in \{a,v\}} \frac{1}{N}  \sum_{i=1}^N D_p ( \bm{\hat{x}_{p_{msk}}}^{(i)})
\end{equation} 
\begin{equation}
\footnotesize
    \mathcal{L}_{adv}^{(D)} = \sum_{p \in \{a,v\}} \frac{1}{N}  \sum_{i=1}^N ( D_p (\bm{\hat{x}_{p_{msk}}}^{(i)}) - D_p (\bm{x_{p_{msk}}}^{(i)}) ) 
\end{equation} 
Here, $D_p$ denotes the discriminator of each modality, and the $\mathcal{L}_{adv}^{(D)}$ and $\mathcal{L}_{adv}^{(G)}$ represents the adversarial loss during the generator and the discriminator training steps respectively. 

The overall training loss for the generative training step is as follows, where the $\lambda_*$ parameters represent the corresponding loss weights:
\begin{align}
\footnotesize
    \mathcal{L}^{(G)} &= \lambda_c \mathcal{L}_c + \lambda_{rec} \mathcal{L}_{rec} + \lambda_{adv} \mathcal{L}_{adv}^{(G)} \label{eq:g_loss}
\end{align}
Computing the autoencoding loss objective on the masked temporal slices strictly enforces the decoder to learn from the other modality, as the input embeddings for the decoder at masked indices are obtained from the other modality. This novel strategy explicitly enforces audio-visual correspondence supplementing the contrastive loss objective. 

\begin{figure}[t]
    \centering
    \includegraphics[width=0.9\linewidth]{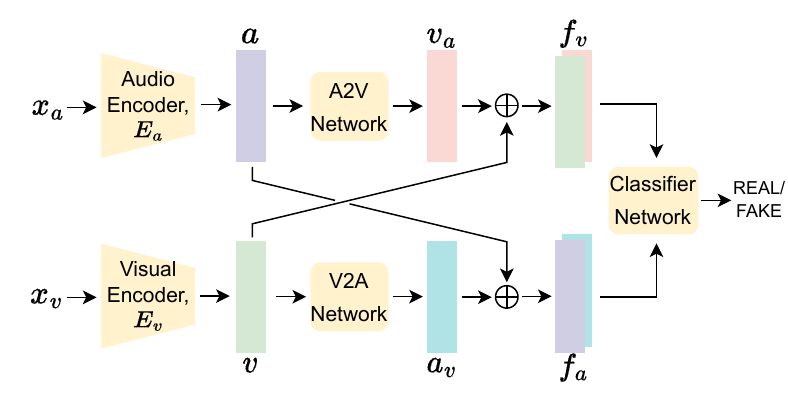}
    \vspace{-0.5em}
    \caption{\textbf{Deepfake Classification Stage.} Given a sample $x\in  \mathcal{D}_{df}$, comprising of audio and visual inputs $\bm{x_a}$ and $\bm{x_v}$, we obtain the unimodal features ($\bm{a}, \bm{v}$) and the cross-modal embeddings ($\bm{a_v}, \bm{v_a}$). For each modality, the unimodal and cross-modal embeddings are concatenated to obtain ($\bm{f_a}, \bm{f_v}$). A classifier network is then trained to take ($\bm{f_a}, \bm{f_v}$) as input and predict if the input is real or fake.}
    \label{fig:pipeline_dfc}
    \vspace{-1em}
\end{figure}

\subsection{Deepfake Classification Stage}
The goal of this stage is to detect video deepfakes, where either or both audio and visual modalities have been faked. 
For this, we use the encoders and the cross-modal networks trained in the representation learning phase.
We train a classifier to tell real videos and deepfakes apart using a supervised learning approach. The classification pipeline is depicted in~\cref{fig:pipeline_dfc}. 
Since the learned representations have a high audio-visual correspondence for real videos, we expect the classifier to exploit the lack of audio-visual cohesion of synthesized samples in distinguishing between real and fake.\looseness=-1

\vspace{.4em}\noindent\textbf{Input Tokenization.}
The process followed in input tokenization is identical to Stage 1 except for the dataset used. In this stage we draw samples from a labeled deepfake dataset ($\mathcal{D}_{df}$) consisting of both real and fake videos, \ie, $(x,y) \in \mathcal{D}_{df}$, where $x$ is the video sample and $y$ is the label (real/fake).\looseness=-1 

\vspace{.4em}\noindent\textbf{Feature Extraction.}
The tokenized inputs ($\bm{x_a}, \bm{x_v}$), are sent through the backbone of Stage 1 to obtain (i) the feature embeddings ($\bm{a}, \bm{v}$), which are the outputs of the uni-modal encoders, and (ii) the cross-modal embeddings ($\bm{a_v}, \bm{v_a}$), which are the outputs of the A2V/V2A cross-modal networks. Here, the cross-modal embeddings are computed for all temporal slices (note: we do not use masking in this stage). We concatenate the two embeddings of each modality creating $(\bm{f_a}, \bm{f_v})$, where
$\bm{f_p} = \bm{p} \, \oplus \,\bm{p_q},\; \forall p,q \in \{a,v\},  p \neq q$,
and $\oplus$ is the concatenation operator along the feature dimension.\looseness=-1

\vspace{.4em}\noindent\textbf{Classifier Network.}
The classifier network, $Q$, takes as input the combined embeddings of each modality $(\bm{f_a}, \bm{f_v})$, and predicts if a given sample is real or fake. The classifier network consists of two uni-modal patch reduction networks: $(\Psi_a, \Psi_v$),  followed by a classifier head, $\Gamma$. Each embedding $(\bm{f_a}, \bm{f_v})$, is first distilled in the patch dimension using the corresponding uni-modal patch reduction networks. Then the output embeddings are concatenated along the feature dimension and fed into the classifier head which outputs the logits, $l$, used to classify if a given sample is real or fake. Formally, $l = Q(\bm{f_a}, \bm{f_v}) = \Gamma( \Psi_a( \bm{f_a} ) \oplus \Psi_v( \bm{f_v} ) ) $.

\vspace{.4em}\noindent\textbf{Loss Function.}
We use the standard cross-entropy loss, denoted by $\mathcal{L}_{CE}$ as the learning objective, computed using the input labels, $y$, and the output logits, $l$.

\vspace{.4em}\noindent\textbf{Deepfake Classifier Inference Stage.}
During inference, we first split the video into blocks of time $T$ (the sample length during training) with a step size of $T/8$, which is the duration of a temporal slice. The output logits are computed for each of the blocks and the classification decision (real or fake) is made based on the mean of the output logits.

\section{Experiments and Results}
\label{sec:results}

\subsection{Implementation}

We train Stage 1 (representation learning), using the LRS3 dataset~\cite{afouras2018lrs3}, which exclusively contains real videos. In Stage 2 (deepfake classification), we train a classifier that follows a supervised learning approach using the FakeAVCeleb~\cite{khalid2021fakeavceleb} dataset. FakeAVCeleb comprises of both real and fake videos, where either one or both audio-visual modalities have been synthesized using different combinations of several generative deepfake algorithms (visual: FaceSwap\cite{korshunova2017faceswap}, FSGAN\cite{nirkin2019fsgan}, and Wav2Lip \cite{prajwal2020wav2lip}; audio: SV2TTS \cite{jia2018sv2tts}). Please refer to the supplementary for additional details on datasets, architecture, and implementation. 

\begin{table}[t]
    \centering
    \begin{adjustbox}{width=0.85\linewidth,center}
    \begin{tabular}{r c *{2}{c}}
    \toprule
    Method & Modality & ACC & AUC   \\
    \midrule
    Xception \cite{rossler2019faceforensics++} & \modv  & 67.9 & 70.5  \\
    LipForensics \cite{haliassos2021lipsdontlie} & \modv & 80.1 & 82.4  \\
    FTCN \cite{zheng2021ftcn} & \modv  & 64.9 & 84.0\\
    CViT \cite{wodajo2021cvit} & \modv & 69.7 & 71.8  \\
    RealForensics \cite{haliassos2022lrealforensics} & \modv & \vul{89.9} & \vul{94.6}\\
    \midrule
    Emotions Don't Lie \cite{mittal2020emotions} & \modav & 78.1 & 79.8  \\
    MDS \cite{chugh2020mds} & \modav & 82.8 & 86.5  \\
    AVFakeNet \cite{ilyas2023avfakenet} & \modav & 78.4 & 83.4  \\
    VFD \cite{cheng2022vfd} & \modav & 81.5 & 86.1 \\
    AVoiD-DF \cite{yang2023avoiddf} & \modav & \avul{83.7} & \avul{89.2}  \\
    \midrule
    \textbf{AVFF (Ours)} & \modav & \textbf{98.6} & \textbf{99.1} \\
    \bottomrule
    \end{tabular}
    \end{adjustbox}
    \caption{\textbf{Intra-Dataset Performance.} We evaluate our method against baselines using a 70\%-30\% train-test split on the FakeAVCeleb dataset, where we achieve state-of-the-art performance by significant margins. Best result is in bold, and second best per modality is underlined.
    }
    \label{tab:intra_eval}
    \vspace{-1em}
\end{table}

\subsection{Evaluation and Discussion}
We evaluate the performance of our model against existing state-of-the-art algorithms,  on multiple criteria: intra-dataset performance, cross-manipulation generalization, and cross-dataset generalization following ~\cite{feng2023avanomaly, yang2023avoiddf}. 
We compare our results against state-of-the-art audio-visual approaches and uni-modal (visual) approaches for completeness. We report accuracy (ACC), average precision (AP), and area under the ROC curve (AUC) averaged across multiple runs with different random seeds. For audio-visual algorithms,  we label a video as fake if either or both audio and visual modalities have been manipulated. To maintain fairness, for uni-modal algorithms we consider a video as fake only if the visual modality has been manipulated. 
Please refer to the supplementary for additional results on robustness to unseen audio and visual perturbations.

\vspace{.4em}\noindent\textbf{Intra-Dataset Performance.}
Following the methodology outlined in \cite{yang2023avoiddf}, our training utilizes 70\% of all FakeAVCeleb samples, while the remaining 30\% constitutes the unseen test set. As denoted in \cref{tab:intra_eval}, our approach demonstrates substantial improvements over the existing state-of-the-art, both in audio-visual (AVoiD-DF \cite{yang2023avoiddf}) and uni-modal (RealForensics \cite{haliassos2022lrealforensics}) deepfake detection. Compared to AVoiD-DF, our method achieves an increase in accuracy of 14.9\% (+9.9\% in AUC), and compared to RealForensics the accuracy increases by 8.7\% (+4.5\% AUC).
Overall, the superior performance of audio-visual methods leveraging cross-modal correspondence is evident, outperforming uni-modal approaches that rely on uni-modal artifacts (\ie visual anomalies) introduced by deepfake algorithms. RealForensics, while competitive, discards the audio modality during detection, limiting its applicability exclusively to visual deepfakes. This hinders its practicality as contemporary deepfakes often involve manipulations in both audio and visual modalities.
The enhanced results of both RealForensics and our proposed method highlight the positive impact of employing a pre-training stage for effective representation learning. This observation aligns with findings in other multi-modal representation learning research across diverse downstream tasks \cite{gong2023cavmae, huang2022audiomae, cai2023marlin}.

\begin{table*}[t]
    \centering
    \begin{adjustbox}{width=0.92\linewidth,center}
    \begin{tabular}{r c *{10}{c} *{2}{c}}
    \toprule
    \multirow{2}{*}{Method} & \multirow{2}{*}{Modality} & \multicolumn{2}{c}{RVFA} & \multicolumn{2}{c}{FVRA-WL} & \multicolumn{2}{c}{FVFA-FS} & \multicolumn{2}{c}{FVFA-GAN} & \multicolumn{2}{c}{FVFA-WL} & \multicolumn{2}{c}{AVG-FV} \\ \cmidrule(lr){3-4} \cmidrule(lr){5-6} \cmidrule(lr){7-8} \cmidrule(lr){9-10} \cmidrule(lr){11-12} \cmidrule(lr){13-14} & & AP & AUC & AP & AUC & AP & AUC & AP & AUC & AP & AUC & AP & AUC\\
    \midrule
    Xception \cite{rossler2019faceforensics++} & \modv & - & - & 88.2 & 88.3 & 92.3 & 93.5 & 67.6 & 68.5 & 91.0 & 91.0 & 84.8 & 85.3\\ 
    LipForensics \cite{haliassos2021lipsdontlie} & \modv & - & - & \textbf{97.8} & \vul{97.7} & \vul{99.9} & \vul{99.9} & 61.5 & 68.1 & 98.6 & 98.7 & 89.4 & 91.1 \\
    FTCN \cite{zheng2021ftcn} & \modv & - & - & 96.2 & 97.4 & \textbf{100.} & \textbf{100.} & 77.4 & 78.3 & 95.6 & 96.5 & 92.3 & 93.1 \\
    RealForensics \cite{haliassos2022lrealforensics} & \modv & - & - & 88.8 & 93.0 & 99.3 & 99.1 & \vul{99.8} & \vul{99.8} & 93.4 & 96.7 & \vul{95.3} & \vul{97.1}\\ 
    \midrule
    AV-DFD \cite{zhou2021avdfd} & \modav & \avul{74.9} & 73.3 & \avul{97.0} & 97.4 & 99.6 & 99.7 & 58.4 & 55.4 & \textbf{100.} & \textbf{100.} & 88.8 & 88.1 \\
    AVAD (LRS2) \cite{feng2023avanomaly} & \modav & 62.4 & 71.6 & 93.6 & 93.7 & 95.3 & 95.8 & 94.1 & 94.3 & 93.8 & 94.1 & 94.2 & 94.5\\
    AVAD (LRS3) \cite{feng2023avanomaly} & \modav & 70.7 & \avul{80.5} & 91.1 & 93.0 & 91.0 & 92.3 & 91.6 & 92.7 & 91.4 & 93.1 & 91.3 & 92.8\\
    \midrule
    \textbf{AVFF (Ours)} & \modav & \textbf{93.3} & \textbf{92.4} & 94.8 & \textbf{98.2} & \textbf{100.} & \textbf{100.} & \textbf{99.9} & \textbf{100.} & \avul{99.4} & \avul{99.8} & \textbf{98.5} & \textbf{99.5}\\
    \bottomrule
    \end{tabular}
    \end{adjustbox}
    \caption{\textbf{Cross-Manipulation Generalization on FakeAVCeleb.}
    We evaluate the model's performance by leaving out one category for testing while training on the rest. We consider the following 5 categories in FakeAVCeleb: (i) \textbf{RVFA}: Real Visual - Fake Audio (SV2TTS), (ii) \textbf{FVRA-WL}: Fake Visual - Real Audio (Wav2Lip), (iii) \textbf{FVFA-FS}: Fake Visual - Fake Audio (FaceSwap + Wav2Lip + SV2TTS), (iv) \textbf{FVFA-GAN}: Fake Visual - Fake Audio (FaceSwapGAN + Wav2Lip + SV2TTS), and (v) \textbf{FVFA-WL}: Fake Visual - Fake Audio (Wav2Lip + SV2TTS). The column titles correspond to the category of the test set. AVG-FV corresponds to the average metrics of the categories containing fake visuals. 
    Best result is in bold, and second best is underlined.
    Our method yields consistently high performance across all manipulation methods while yielding state-of-the-art performance for several categories. 
    \looseness=-1
    }
    \label{tab:cross_manip_eval}
    \vspace{-0.75em}
\end{table*}

\vspace{.4em}\noindent\textbf{Cross-Manipulation Generalization.}
In this experiment, we aim to assess the model's performance on samples generated using previously unseen manipulation methods.
The scalability of deepfake detection algorithms to unseen manipulation methods is crucial for adapting to evolving threats, thus ensuring wide applicability across diverse scenarios. 

Similar to \cite{feng2023avanomaly}, we partition the FakeAVCeleb dataset into five categories: RVFA, FVRA-WL, FVFA-FS, FVFA-GAN, and FVFA-WL, based on the algorithms used to generate the deepfakes. Descriptions of these categories are included in the caption of \cref{tab:cross_manip_eval}, for ease of reference. Using these categories, we evaluate the model leaving one category out for testing while training on the remaining categories. Results are reported in \cref{tab:cross_manip_eval}. 
Our method achieves the best performance in almost all cases (and at par with the rest) and notably, yields consistently enhanced performance (AUC > 92+\%, AP > 93+\%) across all categories, while other baselines (Xception \cite{rossler2019faceforensics++}, LipForensics \cite{haliassos2021lipsdontlie}, FTCN \cite{zheng2021ftcn}, AV-DFD \cite{zhou2021avdfd}) fall short in categories FVFA-GAN and RVFA. 
While the performance of AVAD (an unsupervised method) is intriguing, we can notice how other supervised baselines perform better in most cases. 

\begin{table}[t]
    \centering
    \begin{adjustbox}{width=0.75\linewidth,center}
    \begin{tabular}{r c *{2}{c} }
    \toprule
    Method & Modality & AP & AUC \\
    \midrule
    Xception \cite{rossler2019faceforensics++} & \modv& 76.9 & 77.7 \\
    LipForensics \cite{haliassos2021lipsdontlie} & \modv& 89.5 & 86.6 \\
    FTCN \cite{zheng2021ftcn} & \modv& 66.8 & 68.1 \\
    RealForensics \cite{haliassos2022lrealforensics} & \modv& \textbf{95.7} & \vul{93.6} \\
     \midrule
    AV-DFD \cite{zhou2021avdfd} & \modav & 79.6 & 82.1 \\
    AVAD \cite{feng2023avanomaly} & \modav & 87.6 & 86.9 \\
    \midrule
    \textbf{AVFF (Ours)} & \modav & \avul{93.1} & \textbf{95.5}\\
    \bottomrule
    \end{tabular}
    \end{adjustbox}
    \caption{\textbf{Cross-Dataset Generalization on KoDF.} We evaluate our model's performance against baselines by testing the model trained on the FakeAVCeleb dataset, on a subset of the KoDF dataset. Best result is in bold, and second best is underlined.
    }\label{tab:cross_data_manipul}
    \vspace{-1.5em}
\end{table}

\vspace{.4em}\noindent\textbf{Cross-Dataset Generalization.} We also evaluate the adaptability of the model to a different data distribution, by testing on a subset of the KoDF dataset \cite{kwon2021kodf}, following the evaluation protocol established by \cite{feng2023avanomaly}. We report the results in \cref{tab:cross_data_manipul}, where we perform at par with RealForensics. Overall, we observe a performance trend similar to cross-generalization to unseen generative methods. Please refer to the supplementary for additional cross-dataset generalization results on DF-TIMIT \cite{korshunov2018dftimit} and DFDC \cite{dolhansky2020dfdc} datasets.

\begin{figure}[t]
    \centering
    \includegraphics[width=0.9\linewidth]{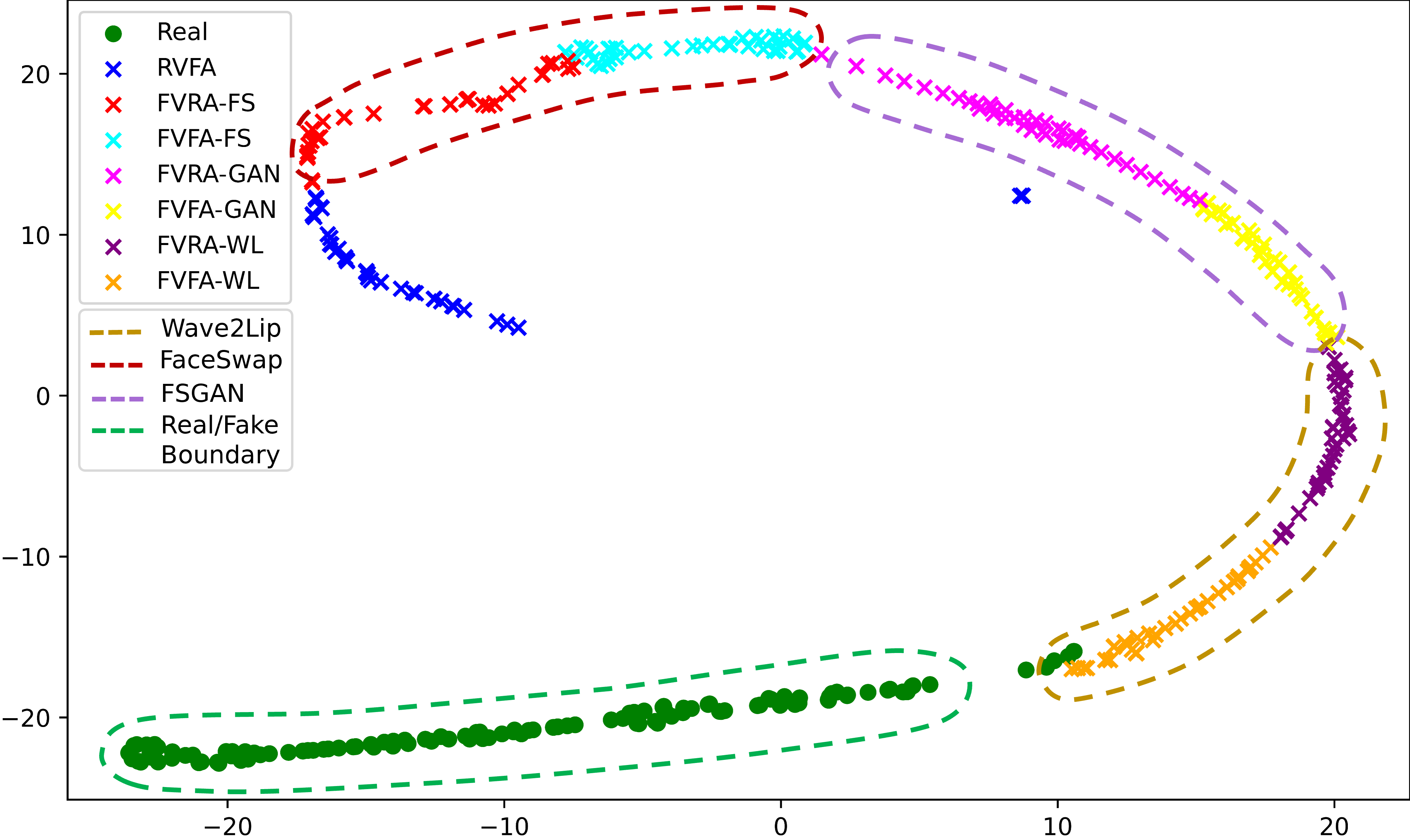}
    \vspace{-0.5em}
    \caption{\textbf{The t-SNE Visualization of the Embeddings at the end of the Representation Learning Stage}. 
    A clear distinction is seen between the representations of real and fake videos, as well as between different deepfake categories. Further analysis indicates that samples of adjacent clusters are generated using the same deepfake algorithm, which we encircle manually to highlight the clusters. \looseness=-1} 
    \label{fig:tsne}
    \vspace{-1.5em}
\end{figure}

\vspace{.4em}\noindent\textbf{Analysis on the Learned Representation.}
During the training of the downstream task, we observe notable AUC scores on the test set, reaching as high as 90\% within the initial 1-3 epochs. Intrigued by this observation, we explore the representations learned at the end of Stage 1 (\cref{subsec:rep_learn}). We visualize the t-SNE plots of embeddings for random samples from each category of the FakeAVCeleb dataset in \cref{fig:tsne}. 
We do not expose Stage 1 to any deepfake samples during training, and still observe clear discrimination between real and fake samples.
This explains the high AUC scores achieved even at the very early stages of the downstream training. A consequence of our representation learning stage is that we observe a clear disentanglement between real and deepfake videos within the FakeAVCeleb dataset. Further, distinct clusters are evident for each deepfake category which indicates that our representations are capable of capturing subtle cues that differentiate different deepfake algorithms despite not encountering any of them during the training stage.
\footnote{Stage 1 is exclusively trained on the LRS3 dataset containing only real videos, while samples for the t-SNE are drawn from FakeAVCeleb.} 
A further analysis of the t-SNE visualizations reveals that the samples belonging to adjacent clusters are related in terms of the deepfake algorithms used to generate them. For instance, FVRA-WL and FVFA-WL, which are adjacent, both employ Wav2Lip to synthesize the deepfakes (refer to the encircled regions in \cref{fig:tsne}). 
These findings underscore the efficacy of our novel audio-visual representation learning paradigm. Please refer to the supplementary for the evaluation of the classification performance on the learned representation which further reinforces the above analysis.

\section{Ablation Study}
\label{sec:ablation}

\noindent\textbf{Autoencoding Objective.} 
In this experiment, we train the model using only the contrastive loss objective, discarding the autoencoding objective. This deactivates complementary masking, cross-modality fusion, and decoding modules. The feature embeddings at the output of the encoders $(\bm{a}, \bm{v})$, are used for the downstream training. Results (see row (i) in \cref{tab:ablations}) indicate a performance reduction, highlighting the importance of the autoencoding objective. 

\vspace{.4em}\noindent\textbf{Cross-Modal Fusion.} 
In this ablation, we discard the A2V/V2A networks, which predict the masked tokens of the other modality, and instead use shared learnable masked tokens similar to MAE approaches \cite{gong2023cavmae, kaiming2022mae, huang2022audiomae, cai2023marlin}. We can see that the performance of the model diminishes (especially AP) (see row (ii) in \cref{tab:ablations}). This highlights the importance of the cross-modal fusion module, as it supplements the representation of a given modality with information extracted from the other modality, helping building the correspondence between the two modalities.

\vspace{.4em}\noindent\textbf{Complementary Masking.} 
Replacing complementary masking with random masking in Stage 1, results in a notable drop in AP and AUC scores, affecting the model's ability to learn correspondences (see row (iii) in \cref{tab:ablations}). We attribute this performance drop to the inability of the model to learn correspondences between audio and visual modalities due to the randomness, which indicates the importance of complementary masking in the proposed method.

\vspace{.4em}\noindent\textbf{Concatenation of Different Embeddings.} 
In the deepfake classification stage, we concatenate the feature embeddings $(\bm{a}, \bm{v})$, with the cross-modal embeddings $(\bm{a_v}, \bm{v_a})$, creating the concatenated embeddings $(\bm{f_a},\bm{f_v})$. In this experiment, we evaluate the performance using each of the embeddings in isolation (see rows (iv) and (v) in \cref{tab:ablations}). While the use of each embedding generates promising results, the synergy of the two embeddings enhances the performance. 

\vspace{.4em}\noindent\textbf{Uni-Modal Patch Reduction.} 
Replacing the uni-modal patch reduction networks ($\Psi_a, \Psi_v$) with Mean Pooling slightly dents the performance  (see row (vi) in \cref{tab:ablations}). {This suggests that a simple linear averaging is sub-optimal in that some subtle cues may no longer be preserved when compared to using an MLP which effectively performs weighted non-linear averaging.}

\begin{table}[t]
    \centering
    \begin{adjustbox}{width=0.9\columnwidth,center}
    \begin{tabular}{l l  *{3}{c}}
    \toprule
    & Method & AP & AUC   \\
    \midrule
    (i) & Only contrastive loss & 84.2 & 90.3 \\
    (ii) & Ours w/o cross-modality fusion & 87.2 & 93.1 \\
    (iii) & Ours w/o complementary masking & 78.9 & 90.7 \\
    (iv) & Only feature embeddings & 89.7 & 97.6 \\
    (v)  & Only cross-modal embeddings & 94.6 & 98.0 \\
    (vi) & Mean Pooling patch dimension & \underline{96.5} & \underline{98.1} \\
    \midrule
    & \textbf{AVFF (Ours)} & \textbf{96.7} & \textbf{99.1} \\    
    \bottomrule
    
    \end{tabular}
    \end{adjustbox}
    \vspace{-0.5em}
    \caption{\textbf{Evaluation on Ablations}. Best result is in bold, and second best is underlined. The proposed AVFF pipeline performs the best among the considered ablations. } \label{tab:ablations}
    \vspace{-1.2em}
\end{table}

\section{Conclusion, Limitations, and Future Work}

In this paper, we propose AVFF, a novel two-staged learning framework for audio-visual deepfake detection. AVFF is composed of a self-supervised representation learning stage that captures the audio-visual correspondence using contrastive learning and a novel complementary masking and cross-modal fusion module within the autoencoding objective, followed by a supervised deepfake classification stage. Our method shows significant improvements in both in-distribution performance as well as generalization on unseen manipulations over both visual-only and audio-visual state-of-the-art algorithms.
Our results not only validate the effectiveness of the proposed approach but also emphasize the potential as a defensive tool against the escalating threat posed by deepfake videos.

\vspace{.4em}\noindent\textbf{Limitations.}
Our work is limited to cases where there is a coherence between the audio and visual modalities, as we rely on it to distinguish fake from real videos. Hence, asynchronous videos with audio lags or mismatched audio, videos with multiple speakers, and visuals with occlusions (\eg, masks, hands) would be challenging. Our model also requires the input to contain both audio and visual modalities,~\ie, videos with only one modality are not supported.

\vspace{.4em}\noindent\textbf{Future Works.} 
A few possible future research avenues include: (i) supplementing our representation learning method with a strategy to preserve uni-modal cues to mitigate the aforementioned limitations, (ii) adaptation for other audio-visual downstream tasks such as emotion recognition, and (iii) generalization for non-humanoid audio-driven videos especially with diffusion-based approaches~\cite{ruan2023mm,jeong2023power}.

\small {
\vspace{.4em}\noindent\textbf{Acknowledgements.}  Yaser Yacoob is supported in part by the DARPA SemaFor Program under HR001120C0124.}

\label{sec:conclusion}

\clearpage
\maketitlesupplementary
\setcounter{section}{0}
\renewcommand{\thesection}{\Alph{section}}

\section{Overview}
This document is structured as follows:
\begin{itemize}
    \item \cref{supp_sec:implementation}:  Implementation details
    \item \cref{supp_sec:datasets}: Dataset details
    \item \cref{supp_sec:results}: Additional results
    \item \cref{supp_sec:limitations}: Visual examples of challenging scenarios
\end{itemize}

\section{Implementation Details} \label{supp_sec:implementation}

\subsection{Representation Learning Stage}
\noindent\textbf{Inputs.}
We draw samples from the LRS3 dataset \cite{afouras2018lrs3}, which exclusively contains real videos. We preprocess all videos as explained in Sec. 3.1 in the main paper. 
The audio stream is converted to a Mel-spectrogram of 128 Mel-frequency bins, with a 16 ms Hamming window every 4 ms.
We randomly sample video clips of $T=3.2$s in duration, sampling 16 visual frames and 768 audio frames (Mel) with clipping/padding where necessary. The 16 visual frames are uniformly sampled such that they are at the first and third quartile of a temporal slice (2 frames/slice $\times$ 8 slices). The visual frames are resized to $224\times224$ spatially and are augmented using random grayscaling and horizontal flipping, each with a probability of 0.5. We make sure that in a given batch, for each sample we draw another sample from the same video but at a different time interval to make sure the model is exposed to the notion of temporal shifts when computing the contrastive loss. Both audio and visual modalities are normalized.

\vspace{.5em}\noindent\textbf{Architecture.}
We adopt the encoder and decoder architectures of each modality from the VideoMAE \cite{tong2022videomae} based on ViT-B \cite{dosovitskiy2020vit}. 
Each of the A2V/V2A networks is composed of a linear layer to match the number of tokens of the other modality followed by a single transformer block. 

\vspace{.5em}\noindent\textbf{Optimization.}
We initialize the audio encoder and decoder using the checkpoint of AudioMAE \cite{huang2022audiomae} pretrained on AudioSet-2M \cite{gemmeke2017audio} and the visual encoder and decoder using the checkpoint of MARLIN \cite{cai2023marlin} pretrained on the YouTubeFace \cite{wolf2011ytfaces} dataset.  
Subsequently, we train the representation learning framework end-to-end, using the AdamW optimizer~\cite{loshchilov2018decoupled} with a learning rate of 1.5e-4 with a cosine decay~\cite{loshchilov2016sgdr}. 
The weights of the losses are as follows: $\lambda_c = 0.01$, $\lambda_{rec} = 1.0$, and $\lambda_{adv} = 0.1$, which were chosen empirically and based on previous research \cite{gong2023cavmae, cai2023marlin}.
We train for 500 epochs with a linear warmup for 40 epochs using a batch size of 32 and a gradient accumulation interval of 2. The training was performed on 4 RTX A6000 GPUs for approximately 60 hours.

\subsection{Deepfake Classification Stage}

\noindent\textbf{Inputs.}
We draw samples from FakeAVCeleb~\cite{khalid2021fakeavceleb}, which consists of deepfake videos where either or both audio and visual modalities have been manipulated. 
The preprocessing and sampling strategy is similar to that of Stage 1, except we do not draw an additional sample from the same video clip as we do not use a contrastive learning objective at this stage. We employ weighted sampling to mitigate the issue of class imbalance between real and fake samples.  

\vspace{.5em}\noindent\textbf{Architecture.}
Each of the uni-modal patch reduction networks is a 3-layer MLP, while the classifier head is a 4-layer MLP. We do not make any changes to the representation learning architecture. 

\vspace{.5em}\noindent\textbf{Optimization.}
We initialize the representation learning framework using the pretrained checkpoint obtained from Stage 1.  Subsequently, we train the pipeline end-to-end, using the AdamW optimizer  \cite{loshchilov2018decoupled} with a cosine annealing with warm restarts scheduler \cite{loshchilov2016sgdr} with a maximum learning rate of 1.0e-4  for 50 epochs with a batch size of 32. The training was performed on 4 RTX A6000 GPUs for approximately 10 hours. 

\begin{table}[t]
    \centering
    \begin{adjustbox}{width=\linewidth,center}
    \begin{tabular}{r c *{4}{c} }
    \toprule
    \multirow{2}{*}{Method} & \multirow{2}{*}{Modality} & \multicolumn{2}{c}{DF-TIMIT} & \multicolumn{2}{c}{DFDC} \\ \cmidrule(lr){3-4}  \cmidrule(lr){5-6} & &  AP & AUC & AP & AUC \\
    \midrule
    Xception \cite{rossler2019faceforensics++} & \modv& 86.0 & 90.5 & 68.0 & 67.6 \\
    LipForensics \cite{haliassos2021lipsdontlie} & \modv&  96.7 & 98.4 & 76.8 & 77.4 \\
    FTCN \cite{zheng2021ftcn} & \modv& \textbf{100.} & 99.8 & 70.5 & 71.1 \\
    RealForensics \cite{haliassos2022lrealforensics} & \modv& 99.2 & 99.5 & \vul{82.9} & \vul{83.7} \\
    \midrule
    \textbf{AVFF (Ours)} & \modav & \textbf{100.} & \textbf{100.} & \textbf{87.0}  & \textbf{86.2} \\
    \bottomrule
    \end{tabular}
    \end{adjustbox}
    \caption{\textbf{Cross-Dataset Generalization.} We evaluate our model's performance against baselines by testing the model trained on the FakeAVCeleb dataset, on the DF-TIMIT dataset and a subset of the DFDC dataset. Best result is in bold, and second best is underlined.
    }\label{supp_tab:cross_dataset}
\end{table}

\begin{figure*}[t]
    \centering
    \includegraphics[width=\linewidth]{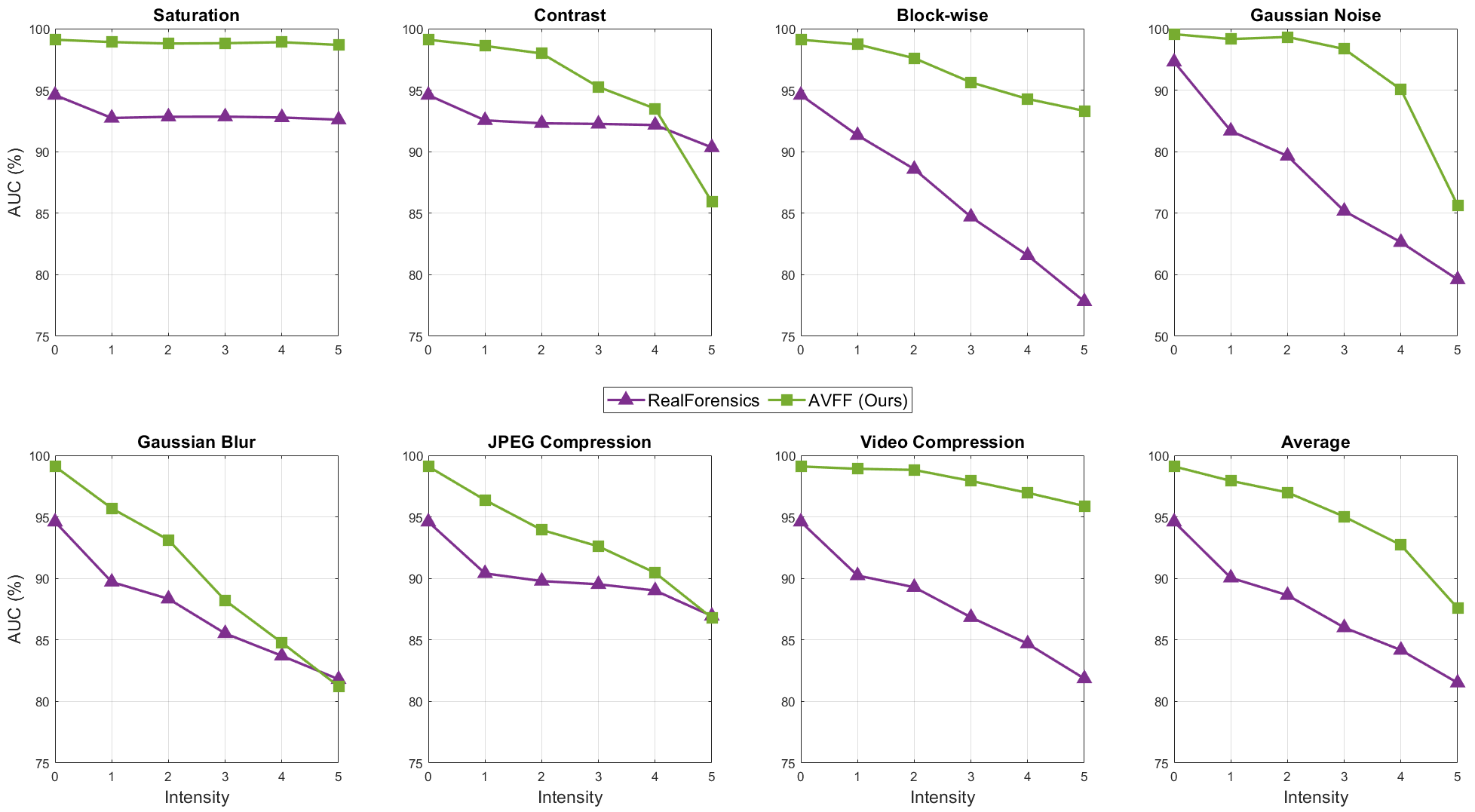}
    \caption{\textbf{Robustness to Unseen Visual Perturbations}. We illustrate AUC scores (\%) as a function of different levels of intensities for various visual perturbations evaluated on the test set of FakeAVCeleb. Our model is more robust than RealForensics \cite{haliassos2022lrealforensics}, which is the current state-of-the-art in robustness to unseen visual perturbations. }
    \label{supp_fig:perturbations}
\end{figure*}

\section{Dataset Details} \label{supp_sec:datasets}

\noindent\textbf{LRS3 \cite{afouras2018lrs3}. } This dataset introduced by Afouras \etal exclusively comprises of real videos. It consists of 5594 videos spanning over 400 hours of TED and TED-X talks in English. The videos in the dataset are processed such that each frame contains faces and the audio and visual streams are in sync.  

\vspace{.5em}\noindent\textbf{FakeAVCeleb \cite{khalid2021fakeavceleb}.} 
The FakeAVCeleb dataset is a deepfake detection dataset, which consists of 20,000 video clips in total. It comprises of 500 real videos sampled from the VoxCeleb2 \cite{chung2018voxceleb2} and 19500 deepfake samples generated using different manipulation methods applied on the set of real videos. The dataset consists of the following manipulations where the deepfake algorithms used in each category are indicated within brackets.  
\begin{itemize}
    \item RVFA: Real Visuals - Fake Audio (SV2TTS \cite{jia2018sv2tts})
    \item FVRA-FS: Fake Visuals - Real Audio (FaceSwap \cite{korshunova2017faceswap})
    \item FVFA-FS: Fake Visuals - Fake Audio (SV2TTS + FaceSwap)
    \item FVFA-GAN: Fake Visuals - Fake Audio (SV2TTS + FaceSwapGAN\cite{nirkin2019fsgan})
    \item FVRA-GAN: Fake Visuals - Real Audio (FaceSwapGAN)
    \item FVRA-WL: Fake Visuals - Real Audio (Wav2Lip \cite{prajwal2020wav2lip})
    \item FVFA-WL: Fake Visuals - Fake Audio (SV2TTS + Wav2Lip)
\end{itemize}

\vspace{.5em}\noindent\textbf{KoDF \cite{kwon2021kodf}.}
This dataset is a large-scale dataset comprising real and synthetic videos of 400+ subjects speaking Korean. KoDF consists of 62K+ real videos and 175K+ fake videos synthesized using the following six algorithms: FaceSwap \cite{korshunova2017faceswap}, DeepFaceLab \cite{perov2020deepfacelab}, FaceSwapGAN\cite{nirkin2019fsgan}, FOMM \cite{siarohin2019fomm}, ATFHP~\cite{yi2020atfhp}, and Wav2Lip \cite{prajwal2020wav2lip}. We use a subset of this dataset following \cite{feng2023avanomaly} to evaluate the cross-dataset generalization performance of our model (Tab. 3 in the main paper).

\vspace{.5em}\noindent\textbf{DF-TIMIT \cite{korshunov2018dftimit}.}
The Deepfake TIMIT dataset comprises deepfake videos manipulated using FaceSwapGAN \cite{nirkin2019fsgan}. The real videos used for manipulation have been sourced by sampling similar-looking identities from the VidTIMIT \cite{sanderson2009vidtimit} dataset. We use their higher-quality (HQ) version, which consists of 320 videos, in evaluating cross-dataset generalization performance. 

\vspace{.5em}\noindent\textbf{DFDC \cite{dolhansky2020dfdc}. } 
The DeepFake Detection Challenge (DFDC) dataset is another deepfake dataset that consists of samples with fake audio besides FakeAVCeleb. It consists of over 100K video clips in total generated using deepfake algorithms such as MM/NN Face Swap \cite{huang2012mmnnfs}, NTH \cite{zakharov2019nth}, FaceSwapGAN \cite{nirkin2019fsgan}, StyleGAN \cite{karras2019stylegan}, and TTS Skins \cite{polyak2019tts}. We use a subset of this dataset consisting of 3215 videos, as used in \cite{haliassos2022lrealforensics, haliassos2021lipsdontlie} to evaluate the model's cross-dataset generalization performance.

\section{Additional Results} \label{supp_sec:results}
In extending our analysis beyond the results outlined in Sec. 4 of the main paper, we conducted additional experiments to provide a more comprehensive evaluation of our model's performance. This supplementary investigation aims to enhance our understanding and confidence in the efficacy of the proposed approach.

\subsection{Cross-Dataset Generalization}
In addition to the cross-dataset generalization evaluation reported on the KoDF dataset (Tab. 3 of the main paper), we further evaluate cross-dataset generalization on the DF-TIMIT dataset and a subset of the DFDC dataset following \cite{haliassos2021lipsdontlie, haliassos2022lrealforensics}.
We are limited to comparing against baselines with open-source codes. We were unable to obtain the models nor results for the other baselines from the authors.   
As illustrated in \cref{supp_tab:cross_dataset}, we achieve the best cross-dataset generalization performance, when evaluated on both DF-TIMIT and DFDC. 

\subsection{Robustness to Unseen Perturbations}
In real-world scenarios, videos undergo post-processing (\eg when sharing through social media platforms), which perturbs both audio and visual modalities. Hence, it is crucial for a model to be robust to unseen perturbations. To this end, we evaluate the performance of our model (trained without augmentations) on several unseen perturbations applied to each modality.

\begin{figure}[t]
    \centering
    \includegraphics[width=\linewidth]{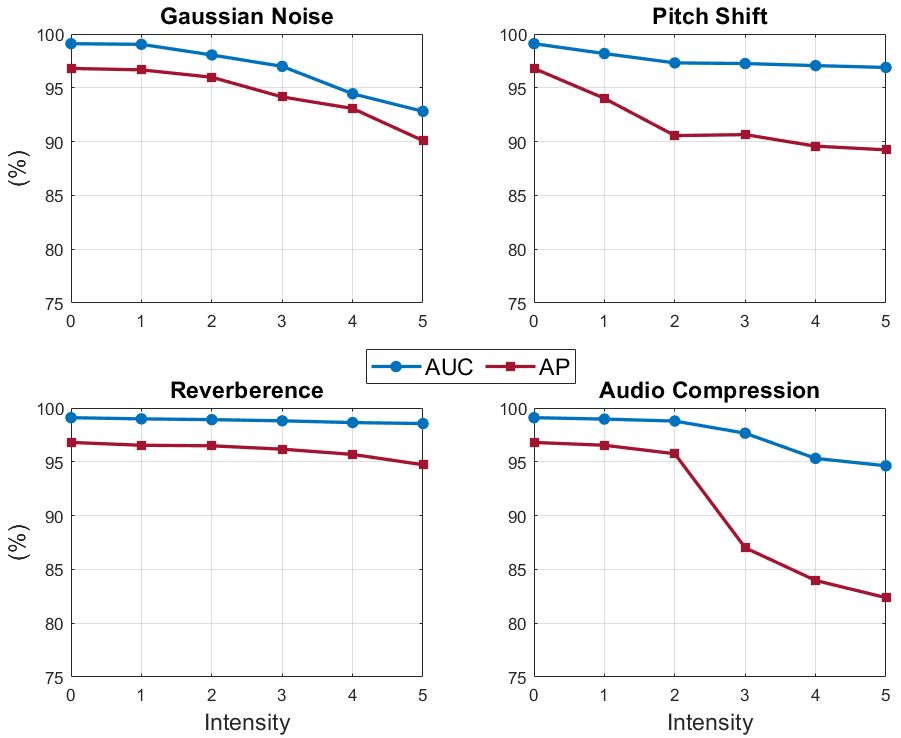}
    \caption{\textbf{Robustness to Unseen Audio Perturbations}. We illustrate the variation of AUC and AP scores (\%) as a function of different levels of intensities for various audio perturbations evaluated on the test set of FakeAVCeleb. Overall our model depicts impressive robustness to audio perturbations.}
    \label{supp_fig:perturbations_aud}
\end{figure}

\vspace{.5em}\noindent\textbf{Visual Perturbations.}
Following \cite{haliassos2022lrealforensics, feng2023avanomaly, haliassos2021lipsdontlie}, we evaluate the performance on the following perturbations: saturation, contrast, block-wise distortion, Gaussian noise, Gaussian blur, JPEG compression, and video compression on five different levels of intensities. The implementations for the perturbations and the levels of intensities were sourced from the official repository of DeeperForensics-1.0 \cite{jiang2020deeperforensics}. 
We compare our model's performance against RealForensics \cite{haliassos2022lrealforensics}, which has the current state-of-the-art performance in robustness to unseen visual perturbations \cite{haliassos2022lrealforensics, feng2023avanomaly}. 
As depicted in \cref{supp_fig:perturbations}, our model demonstrates enhanced robustness against unseen visual perturbations compared to RealForensics in most scenarios. Particularly, noteworthy improvements are observed in cases of block-wise distortion, Gaussian noise, and video compression.

\vspace{.5em}\noindent\textbf{Audio Perturbations.}
In this experiment, we subject the audio stream to a range of perturbations: Gaussian Noise, pitch shift, changes in reverberance, and audio compression. The performance of our model under these perturbations is illustrated across five intensity levels in \cref{supp_fig:perturbations_aud}. 
To generate the perturbed audio samples, we employ the following Python libraries: \textit{torchaudio} (Gaussian noise, pitch shift), \textit{pysndfx} (reverberance), and \textit{pydub} (audio compression). 
The parameters used to generate samples at each intensity level are tabulated in \cref{supp_tab:param_aud_pert}. 
As seen in \cref{supp_fig:perturbations_aud}, overall our model is robust to various audio perturbations. 
A slight decrease in performance is seen in cases of Gaussian noise and pitch shift, with the increase in intensity. 
Notably, the model showcases high robustness to changes in reverberance, with minimal fluctuations across all intensity levels.
However, a noticeable reduction in average precision is observed for high-intensity levels of audio compression, potentially due to artifacts introduced by the reduced bitrate in extreme compression scenarios.

\begin{table}[t]
    \centering
    \begin{adjustbox}{width=\linewidth,center}
    \begin{tabular}{l *{5}{r} }
    \toprule
    \multirow{2}{*}{Perturbation} & \multicolumn{5}{c}{Intensity Level} \\ \cmidrule(lr){2-6}  &  1 & 2 & 3 & 4 & 5 \\
    \midrule
    Gaussian Noise (SNR) & 40 & 30 & 20 & 15 & 10 \\
    Pitch Shift (steps) & $\pm$2 & $\pm$4 & $\pm$6 & $\pm$8 & $\pm$10 \\
    Reverberance & 20 & 40 & 60 & 80 & 100 \\
    Audio Compression (bitrate) & 320k & 256k & 192k & 128k & 64k \\  
    \bottomrule
    \end{tabular}
    \end{adjustbox}
    \caption{Parameters used to generate samples with audio perturbations at different levels of intensities.}
    \label{supp_tab:param_aud_pert}
\end{table}

\begin{table}[t]
    \centering
    \begin{adjustbox}{width=0.9\columnwidth,center}
    \begin{tabular}{l l  *{3}{c}}
    \toprule
    & Method & ACC & AUC   \\
    \midrule
    (i) & Ours with Frozen Stage 1 + MLP & 94.8 & 85.3 \\
    (ii) & Ours with Frozen Stage 1 + SVM & 96.3 & 88.9 \\
    \midrule
    & \textbf{AVFF (Ours)} & \textbf{98.6} & \textbf{99.1} \\    
    \bottomrule
    
    \end{tabular}
    \end{adjustbox}
    \caption{\textbf{Classification Performance on the Learned Representation.} We evaluate the classification performance of the learned representation by freezing the encoders and A2V/V2A networks and training only the downstream networks. We employ (i) an MLP similar to the proposed method, and (ii) a kernel SVM (RBF), as the classifier. Both classifiers yield reasonably high metrics, indicating the effectiveness of the learned representation at the end of Stage 1 in distinguishing between real and fake videos.
    } \label{supp_tab:classification}
\end{table}

\subsection{Classification Performance on the Learned Representation}

We further evaluate the learned representation at the end of Stage 1, by performing the downstream deepfake classification task with the weights of the encoders and the A2V/V2A networks frozen. 
We train two classifiers for the downstream task: (i) the classifier network described in the main paper, and (ii) kernel SVM using an RBF kernel (gamma=0.1, C=1.0).   
The results are reported in \cref{supp_tab:classification}. Both classifiers yield reasonably high accuracy and AUC values. This is indicative of the highly discriminative nature of the learned representation at the end of Stage 1.
This reinforces the analysis on the learned representation at the end of Stage 1, as discussed in Sec. 4.2 of the main paper.

\section{Visual Examples of Challenging Scenarios} \label{supp_sec:limitations}

\begin{figure}[t]
    \centering
    \includegraphics[width=\linewidth]{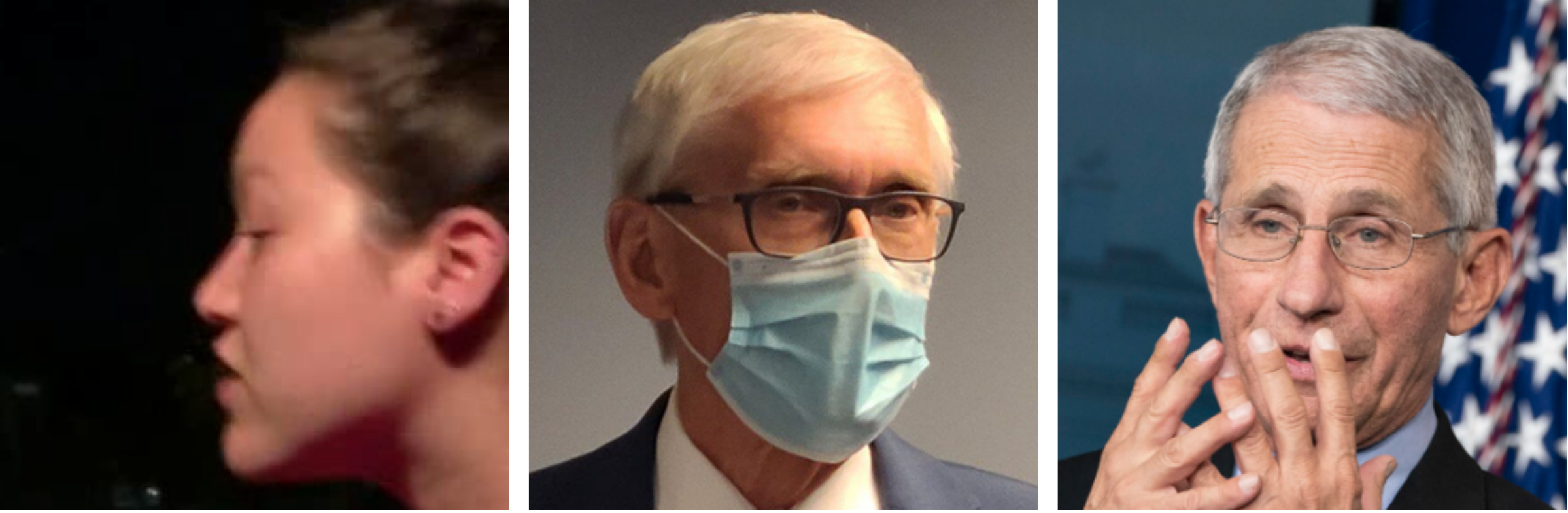}
    \caption{\textbf{Visual Examples of a Few Challenging Scenarios}. Images from left to right depict examples of extreme poses (\eg near profile), occlusions with masks, and occlusions with hands across the face, which makes it challenging for our model to establish correspondence between the audio and visual modalities.}
    \label{supp_fig:limitations}
\end{figure}

As discussed in Sec. 6 of the main paper, since we rely on audio-visual correspondence to distinguish between real and fake videos, scenarios where such correspondence cannot be established would be challenging. 
In \cref{supp_fig:limitations}, we depict a few such visual examples.

\pagebreak

{
    \small
    \bibliographystyle{ieeenat_fullname}
    \bibliography{main}

\begin{thebibliography}{74}
\providecommand{\natexlab}[1]{#1}
\providecommand{\url}[1]{\texttt{#1}}
\expandafter\ifx\csname urlstyle\endcsname\relax
  \providecommand{\doi}[1]{doi: #1}\else
  \providecommand{\doi}{doi: \begingroup \urlstyle{rm}\Url}\fi

\bibitem[Afouras et~al.(2018)Afouras, Chung, and Zisserman]{afouras2018lrs3}
Triantafyllos Afouras, Joon~Son Chung, and Andrew Zisserman.
\newblock Lrs3-ted: a large-scale dataset for visual speech recognition.
\newblock \emph{arXiv preprint arXiv:1809.00496}, 2018.

\bibitem[Agarwal et~al.(2020)Agarwal, Farid, Fried, and Agrawala]{agarwal2020detecting}
Shruti Agarwal, Hany Farid, Ohad Fried, and Maneesh Agrawala.
\newblock Detecting deep-fake videos from phoneme-viseme mismatches.
\newblock In \emph{Proceedings of the IEEE/CVF conference on computer vision and pattern recognition workshops}, pages 660--661, 2020.

\bibitem[Arjovsky et~al.(2017)Arjovsky, Chintala, and Bottou]{arjovsky2017wasserstein}
Martin Arjovsky, Soumith Chintala, and L{\'e}on Bottou.
\newblock Wasserstein generative adversarial networks.
\newblock In \emph{International conference on machine learning}, pages 214--223. PMLR, 2017.

\bibitem[Bai et~al.(2023)Bai, Liu, Zhang, Li, and Hu]{bai2023aunet}
Weiming Bai, Yufan Liu, Zhipeng Zhang, Bing Li, and Weiming Hu.
\newblock Aunet: Learning relations between action units for face forgery detection.
\newblock In \emph{2023 IEEE/CVF Conference on Computer Vision and Pattern Recognition (CVPR)}, pages 24709--24719, 2023.

\bibitem[Baltrušaitis et~al.(2019)Baltrušaitis, Ahuja, and Morency]{baltrusaitis2019multimodal}
Tadas Baltrušaitis, Chaitanya Ahuja, and Louis-Philippe Morency.
\newblock Multimodal machine learning: A survey and taxonomy.
\newblock \emph{IEEE Transactions on Pattern Analysis and Machine Intelligence}, 41\penalty0 (2):\penalty0 423--443, 2019.

\bibitem[Cai et~al.(2023)Cai, Ghosh, Stefanov, Dhall, Cai, Rezatofighi, Haffari, and Hayat]{cai2023marlin}
Zhixi Cai, Shreya Ghosh, Kalin Stefanov, Abhinav Dhall, Jianfei Cai, Hamid Rezatofighi, Reza Haffari, and Munawar Hayat.
\newblock Marlin: Masked autoencoder for facial video representation learning.
\newblock In \emph{Proceedings of the IEEE/CVF Conference on Computer Vision and Pattern Recognition}, pages 1493--1504, 2023.

\bibitem[Cheng et~al.(2022)Cheng, Guo, Wang, Li, Chang, and Nie]{cheng2022vfd}
Harry Cheng, Yangyang Guo, Tianyi Wang, Qi Li, Xiaojun Chang, and Liqiang Nie.
\newblock Voice-face homogeneity tells deepfake.
\newblock \emph{arXiv preprint arXiv:2203.02195}, 2022.

\bibitem[Chugh et~al.(2020)Chugh, Gupta, Dhall, and Subramanian]{chugh2020mds}
Komal Chugh, Parul Gupta, Abhinav Dhall, and Ramanathan Subramanian.
\newblock Not made for each other-audio-visual dissonance-based deepfake detection and localization.
\newblock In \emph{Proceedings of the 28th ACM international conference on multimedia}, pages 439--447, 2020.

\bibitem[Chung et~al.(2018)Chung, Nagrani, and Zisserman]{chung2018voxceleb2}
J Chung, A Nagrani, and A Zisserman.
\newblock Voxceleb2: Deep speaker recognition.
\newblock \emph{Interspeech 2018}, 2018.

\bibitem[Chung and Zisserman(2016)]{Chung16a-syncnet}
J.~S. Chung and A. Zisserman.
\newblock Out of time: automated lip sync in the wild.
\newblock In \emph{Workshop on Multi-view Lip-reading, ACCV}, 2016.

\bibitem[Devlin et~al.(2019)Devlin, Chang, Lee, and Toutanova]{Devlin2019BERTPO}
Jacob Devlin, Ming-Wei Chang, Kenton Lee, and Kristina Toutanova.
\newblock Bert: Pre-training of deep bidirectional transformers for language understanding.
\newblock In \emph{North American Chapter of the Association for Computational Linguistics}, 2019.

\bibitem[Dolhansky et~al.(2020)Dolhansky, Bitton, Pflaum, Lu, Howes, Wang, and Ferrer]{dolhansky2020dfdc}
Brian Dolhansky, Joanna Bitton, Ben Pflaum, Jikuo Lu, Russ Howes, Menglin Wang, and Cristian~Canton Ferrer.
\newblock The deepfake detection challenge (dfdc) dataset.
\newblock \emph{arXiv preprint arXiv:2006.07397}, 2020.

\bibitem[Dong et~al.(2023)Dong, Wang, Ji, Liang, Fan, and Ge]{Dong_2023_CVPR}
Shichao Dong, Jin Wang, Renhe Ji, Jiajun Liang, Haoqiang Fan, and Zheng Ge.
\newblock Implicit identity leakage: The stumbling block to improving deepfake detection generalization.
\newblock In \emph{Proceedings of the IEEE/CVF Conference on Computer Vision and Pattern Recognition (CVPR)}, pages 3994--4004, 2023.

\bibitem[Dong et~al.(2022)Dong, Bao, Chen, Zhang, Zhang, Yu, Chen, Wen, and Guo]{Dong2022protecting}
Xiaoyi Dong, Jianmin Bao, Dongdong Chen, Ting Zhang, Weiming Zhang, Nenghai Yu, Dong Chen, Fang Wen, and Baining Guo.
\newblock Protecting celebrities from deepfake with identity consistency transformer.
\newblock In \emph{2022 IEEE/CVF Conference on Computer Vision and Pattern Recognition (CVPR)}, pages 9458--9468, 2022.

\bibitem[Dosovitskiy et~al.(2020)Dosovitskiy, Beyer, Kolesnikov, Weissenborn, Zhai, Unterthiner, Dehghani, Minderer, Heigold, Gelly, et~al.]{dosovitskiy2020vit}
Alexey Dosovitskiy, Lucas Beyer, Alexander Kolesnikov, Dirk Weissenborn, Xiaohua Zhai, Thomas Unterthiner, Mostafa Dehghani, Matthias Minderer, Georg Heigold, Sylvain Gelly, et~al.
\newblock An image is worth 16x16 words: Transformers for image recognition at scale.
\newblock In \emph{International Conference on Learning Representations}, 2020.

\bibitem[Feng et~al.(2023)Feng, Chen, and Owens]{feng2023avanomaly}
Chao Feng, Ziyang Chen, and Andrew Owens.
\newblock Self-supervised video forensics by audio-visual anomaly detection.
\newblock In \emph{Proceedings of the IEEE/CVF Conference on Computer Vision and Pattern Recognition}, pages 10491--10503, 2023.

\bibitem[Gemmeke et~al.(2017)Gemmeke, Ellis, Freedman, Jansen, Lawrence, Moore, Plakal, and Ritter]{gemmeke2017audio}
Jort~F Gemmeke, Daniel~PW Ellis, Dylan Freedman, Aren Jansen, Wade Lawrence, R~Channing Moore, Manoj Plakal, and Marvin Ritter.
\newblock Audio set: An ontology and human-labeled dataset for audio events.
\newblock In \emph{2017 IEEE international conference on acoustics, speech and signal processing (ICASSP)}, pages 776--780. IEEE, 2017.

\bibitem[Georgescu et~al.(2023)Georgescu, Fonseca, Ionescu, Lucic, Schmid, and Arnab]{georgescu2023avmae}
Mariana-Iuliana Georgescu, Eduardo Fonseca, Radu~Tudor Ionescu, Mario Lucic, Cordelia Schmid, and Anurag Arnab.
\newblock Audiovisual masked autoencoders.
\newblock In \emph{Proceedings of the IEEE/CVF International Conference on Computer Vision}, pages 16144--16154, 2023.

\bibitem[Gong et~al.(2023)Gong, Rouditchenko, Liu, Harwath, Karlinsky, Kuehne, and Glass]{gong2023cavmae}
Yuan Gong, Andrew Rouditchenko, Alexander~H. Liu, David Harwath, Leonid Karlinsky, Hilde Kuehne, and James~R. Glass.
\newblock Contrastive audio-visual masked autoencoder.
\newblock In \emph{{ICLR}}. OpenReview.net, 2023.

\bibitem[Guzhov et~al.(2022)Guzhov, Raue, Hees, and Dengel]{guzhov2022audioclip}
Andrey Guzhov, Federico Raue, J{\"o}rn Hees, and Andreas Dengel.
\newblock Audioclip: Extending clip to image, text and audio.
\newblock In \emph{ICASSP 2022-2022 IEEE International Conference on Acoustics, Speech and Signal Processing (ICASSP)}, pages 976--980. IEEE, 2022.

\bibitem[Haliassos et~al.(2021)Haliassos, Vougioukas, Petridis, and Pantic]{haliassos2021lipsdontlie}
Alexandros Haliassos, Konstantinos Vougioukas, Stavros Petridis, and Maja Pantic.
\newblock Lips don't lie: A generalisable and robust approach to face forgery detection.
\newblock In \emph{Proceedings of the IEEE/CVF conference on computer vision and pattern recognition}, pages 5039--5049, 2021.

\bibitem[Haliassos et~al.(2022)Haliassos, Mira, Petridis, and Pantic]{haliassos2022lrealforensics}
Alexandros Haliassos, Rodrigo Mira, Stavros Petridis, and Maja Pantic.
\newblock Leveraging real talking faces via self-supervision for robust forgery detection.
\newblock In \emph{Proceedings of the IEEE/CVF Conference on Computer Vision and Pattern Recognition}, pages 14950--14962, 2022.

\bibitem[He et~al.(2022)He, Chen, Xie, Li, Dollár, and Girshick]{kaiming2022mae}
Kaiming He, Xinlei Chen, Saining Xie, Yanghao Li, Piotr Dollár, and Ross Girshick.
\newblock Masked autoencoders are scalable vision learners.
\newblock In \emph{2022 IEEE/CVF Conference on Computer Vision and Pattern Recognition (CVPR)}, pages 15979--15988, 2022.

\bibitem[Huang et~al.(2023)Huang, Wang, Yang, Ai, Zou, Wang, and Ye]{Huang_2023_CVPR}
Baojin Huang, Zhongyuan Wang, Jifan Yang, Jiaxin Ai, Qin Zou, Qian Wang, and Dengpan Ye.
\newblock Implicit identity driven deepfake face swapping detection.
\newblock In \emph{Proceedings of the IEEE/CVF Conference on Computer Vision and Pattern Recognition (CVPR)}, pages 4490--4499, 2023.

\bibitem[Huang and De~La~Torre(2012)]{huang2012mmnnfs}
Dong Huang and Fernando De~La~Torre.
\newblock Facial action transfer with personalized bilinear regression.
\newblock In \emph{Computer Vision--ECCV 2012: 12th European Conference on Computer Vision, Florence, Italy, October 7-13, 2012, Proceedings, Part II 12}, pages 144--158. Springer, 2012.

\bibitem[Huang et~al.(2022)Huang, Xu, Li, Baevski, Auli, Galuba, Metze, and Feichtenhofer]{huang2022audiomae}
Po-Yao Huang, Hu Xu, Juncheng Li, Alexei Baevski, Michael Auli, Wojciech Galuba, Florian Metze, and Christoph Feichtenhofer.
\newblock Masked autoencoders that listen.
\newblock \emph{Advances in Neural Information Processing Systems}, 35:\penalty0 28708--28720, 2022.

\bibitem[Ilyas et~al.(2023)Ilyas, Javed, and Malik]{ilyas2023avfakenet}
Hafsa Ilyas, Ali Javed, and Khalid~Mahmood Malik.
\newblock Avfakenet: A unified end-to-end dense swin transformer deep learning model for audio--visual deepfakes detection.
\newblock \emph{Applied Soft Computing}, 136:\penalty0 110124, 2023.

\bibitem[Jeong et~al.(2023)Jeong, Ryoo, Lee, Seo, Byeon, Kim, and Kim]{jeong2023power}
Yujin Jeong, Wonjeong Ryoo, Seunghyun Lee, Dabin Seo, Wonmin Byeon, Sangpil Kim, and Jinkyu Kim.
\newblock The power of sound (tpos): Audio reactive video generation with stable diffusion.
\newblock In \emph{Proceedings of the IEEE/CVF International Conference on Computer Vision}, pages 7822--7832, 2023.

\bibitem[Jia et~al.(2018)Jia, Zhang, Weiss, Wang, Shen, Ren, Nguyen, Pang, Lopez~Moreno, Wu, et~al.]{jia2018sv2tts}
Ye Jia, Yu Zhang, Ron Weiss, Quan Wang, Jonathan Shen, Fei Ren, Patrick Nguyen, Ruoming Pang, Ignacio Lopez~Moreno, Yonghui Wu, et~al.
\newblock Transfer learning from speaker verification to multispeaker text-to-speech synthesis.
\newblock \emph{Advances in neural information processing systems}, 31, 2018.

\bibitem[Jiang et~al.(2020)Jiang, Li, Wu, Qian, and Loy]{jiang2020deeperforensics}
Liming Jiang, Ren Li, Wayne Wu, Chen Qian, and Chen~Change Loy.
\newblock Deeperforensics-1.0: A large-scale dataset for real-world face forgery detection.
\newblock In \emph{Proceedings of the IEEE/CVF conference on computer vision and pattern recognition}, pages 2889--2898, 2020.

\bibitem[Jung et~al.(2020)Jung, Kim, and Kim]{Jung2020eye}
Tackhyun Jung, Sangwon Kim, and Keecheon Kim.
\newblock Deepvision: Deepfakes detection using human eye blinking pattern.
\newblock \emph{IEEE Access}, 8:\penalty0 83144--83154, 2020.

\bibitem[Kaddar et~al.(2021)Kaddar, Fezza, Hamidouche, Akhtar, and Hadid]{kaddar2021hcit}
Bachir Kaddar, Sid~Ahmed Fezza, Wassim Hamidouche, Zahid Akhtar, and Abdenour Hadid.
\newblock Hcit: Deepfake video detection using a hybrid model of cnn features and vision transformer.
\newblock In \emph{2021 International Conference on Visual Communications and Image Processing (VCIP)}, pages 1--5, 2021.

\bibitem[Karras et~al.(2019)Karras, Laine, and Aila]{karras2019stylegan}
Tero Karras, Samuli Laine, and Timo Aila.
\newblock A style-based generator architecture for generative adversarial networks.
\newblock In \emph{Proceedings of the IEEE/CVF conference on computer vision and pattern recognition}, pages 4401--4410, 2019.

\bibitem[Khalid et~al.(2021)Khalid, Tariq, Kim, and Woo]{khalid2021fakeavceleb}
Hasam Khalid, Shahroz Tariq, Minha Kim, and Simon~S Woo.
\newblock Fakeavceleb: A novel audio-video multimodal deepfake dataset.
\newblock \emph{arXiv preprint arXiv:2108.05080}, 2021.

\bibitem[Korshunov and Marcel(2018)]{korshunov2018dftimit}
Pavel Korshunov and S{\'e}bastien Marcel.
\newblock Deepfakes: a new threat to face recognition? assessment and detection.
\newblock \emph{arXiv preprint arXiv:1812.08685}, 2018.

\bibitem[Korshunova et~al.(2017)Korshunova, Shi, Dambre, and Theis]{korshunova2017faceswap}
Iryna Korshunova, Wenzhe Shi, Joni Dambre, and Lucas Theis.
\newblock Fast face-swap using convolutional neural networks.
\newblock In \emph{Proceedings of the IEEE international conference on computer vision}, pages 3677--3685, 2017.

\bibitem[Kwon et~al.(2021)Kwon, You, Nam, Park, and Chae]{kwon2021kodf}
Patrick Kwon, Jaeseong You, Gyuhyeon Nam, Sungwoo Park, and Gyeongsu Chae.
\newblock Kodf: A large-scale korean deepfake detection dataset.
\newblock In \emph{Proceedings of the IEEE/CVF International Conference on Computer Vision}, pages 10744--10753, 2021.

\bibitem[Li et~al.(2018)Li, Chang, and Lyu]{Li2018eye}
Yuezun Li, Ming-Ching Chang, and Siwei Lyu.
\newblock In ictu oculi: Exposing ai created fake videos by detecting eye blinking.
\newblock In \emph{2018 IEEE International Workshop on Information Forensics and Security (WIFS)}, pages 1--7, 2018.

\bibitem[Loshchilov and Hutter(2016)]{loshchilov2016sgdr}
Ilya Loshchilov and Frank Hutter.
\newblock Sgdr: Stochastic gradient descent with warm restarts.
\newblock In \emph{International Conference on Learning Representations}, 2016.

\bibitem[Loshchilov and Hutter(2018)]{loshchilov2018decoupled}
Ilya Loshchilov and Frank Hutter.
\newblock Decoupled weight decay regularization.
\newblock In \emph{International Conference on Learning Representations}, 2018.

\bibitem[Lutz and Bassett(2021)]{lutz2021head}
Kevin Lutz and Robert Bassett.
\newblock Deepfake detection with inconsistent head poses: Reproducibility and analysis.
\newblock \emph{CoRR}, abs/2108.12715, 2021.

\bibitem[Ma et~al.(2021)Ma, Simon, Saragih, Wang, Li, De~la Torre, and Sheikh]{Ma_2021_CVPR}
Shugao Ma, Tomas Simon, Jason Saragih, Dawei Wang, Yuecheng Li, Fernando De~la Torre, and Yaser Sheikh.
\newblock Pixel codec avatars.
\newblock In \emph{Proceedings of the IEEE/CVF Conference on Computer Vision and Pattern Recognition (CVPR)}, pages 64--73, 2021.

\bibitem[Mittal et~al.(2020{\natexlab{a}})Mittal, Bhattacharya, Chandra, Bera, and Manocha]{mittal2020emotions}
Trisha Mittal, Uttaran Bhattacharya, Rohan Chandra, Aniket Bera, and Dinesh Manocha.
\newblock Emotions don't lie: An audio-visual deepfake detection method using affective cues.
\newblock In \emph{Proceedings of the 28th ACM international conference on multimedia}, pages 2823--2832, 2020{\natexlab{a}}.

\bibitem[Mittal et~al.(2020{\natexlab{b}})Mittal, Bhattacharya, Chandra, Bera, and Manocha]{mittal2020m3er}
Trisha Mittal, Uttaran Bhattacharya, Rohan Chandra, Aniket Bera, and Dinesh Manocha.
\newblock M3er: Multiplicative multimodal emotion recognition using facial, textual, and speech cues.
\newblock In \emph{Proceedings of the AAAI conference on artificial intelligence}, pages 1359--1367, 2020{\natexlab{b}}.

\bibitem[Morgado et~al.(2021)Morgado, Misra, and Vasconcelos]{morgado2021robust}
Pedro Morgado, Ishan Misra, and Nuno Vasconcelos.
\newblock Robust audio-visual instance discrimination.
\newblock In \emph{Proceedings of the IEEE/CVF Conference on Computer Vision and Pattern Recognition}, pages 12934--12945, 2021.

\bibitem[Nirkin et~al.(2019)Nirkin, Keller, and Hassner]{nirkin2019fsgan}
Yuval Nirkin, Yosi Keller, and Tal Hassner.
\newblock Fsgan: Subject agnostic face swapping and reenactment.
\newblock In \emph{Proceedings of the IEEE/CVF international conference on computer vision}, pages 7184--7193, 2019.

\bibitem[Ojha et~al.(2023)Ojha, Li, and Lee]{Ojha_2023_CVPR}
Utkarsh Ojha, Yuheng Li, and Yong~Jae Lee.
\newblock Towards universal fake image detectors that generalize across generative models.
\newblock In \emph{Proceedings of the IEEE/CVF Conference on Computer Vision and Pattern Recognition (CVPR)}, pages 24480--24489, 2023.

\bibitem[Perov et~al.(2020)Perov, Gao, Chervoniy, Liu, Marangonda, Um{\'e}, Dpfks, Facenheim, RP, Jiang, et~al.]{perov2020deepfacelab}
Ivan Perov, Daiheng Gao, Nikolay Chervoniy, Kunlin Liu, Sugasa Marangonda, Chris Um{\'e}, Mr Dpfks, Carl~Shift Facenheim, Luis RP, Jian Jiang, et~al.
\newblock Deepfacelab: Integrated, flexible and extensible face-swapping framework.
\newblock \emph{arXiv preprint arXiv:2005.05535}, 2020.

\bibitem[Polyak et~al.(2019)Polyak, Wolf, and Taigman]{polyak2019tts}
Adam Polyak, Lior Wolf, and Yaniv Taigman.
\newblock Tts skins: Speaker conversion via asr.
\newblock \emph{arXiv preprint arXiv:1904.08983}, 2019.

\bibitem[Prajwal et~al.(2020)Prajwal, Mukhopadhyay, Namboodiri, and Jawahar]{prajwal2020wav2lip}
KR Prajwal, Rudrabha Mukhopadhyay, Vinay~P Namboodiri, and CV Jawahar.
\newblock A lip sync expert is all you need for speech to lip generation in the wild.
\newblock In \emph{Proceedings of the 28th ACM international conference on multimedia}, pages 484--492, 2020.

\bibitem[Radford et~al.(2021)Radford, Kim, Hallacy, Ramesh, Goh, Agarwal, Sastry, Askell, Mishkin, Clark, et~al.]{radford2021clip}
Alec Radford, Jong~Wook Kim, Chris Hallacy, Aditya Ramesh, Gabriel Goh, Sandhini Agarwal, Girish Sastry, Amanda Askell, Pamela Mishkin, Jack Clark, et~al.
\newblock Learning transferable visual models from natural language supervision.
\newblock In \emph{International conference on machine learning}, pages 8748--8763. PMLR, 2021.

\bibitem[Rossler et~al.(2019)Rossler, Cozzolino, Verdoliva, Riess, Thies, and Nie{\ss}ner]{rossler2019faceforensics++}
Andreas Rossler, Davide Cozzolino, Luisa Verdoliva, Christian Riess, Justus Thies, and Matthias Nie{\ss}ner.
\newblock Faceforensics++: Learning to detect manipulated facial images.
\newblock In \emph{Proceedings of the IEEE/CVF international conference on computer vision}, pages 1--11, 2019.

\bibitem[Rouditchenko et~al.(2021)Rouditchenko, Boggust, Harwath, Chen, Joshi, Thomas, Audhkhasi, Kuehne, Panda, Feris, et~al.]{rouditchenko2021avlnet}
Andrew Rouditchenko, Angie Boggust, David Harwath, Brian Chen, Dhiraj Joshi, Samuel Thomas, Kartik Audhkhasi, Hilde Kuehne, Rameswar Panda, Rogerio Feris, et~al.
\newblock Avlnet: Learning audio-visual language representations from instructional videos.
\newblock In \emph{Annual Conference of the International Speech Communication Association}. International Speech Communication Association, 2021.

\bibitem[Ruan et~al.(2023)Ruan, Ma, Yang, He, Liu, Fu, Yuan, Jin, and Guo]{ruan2023mm}
Ludan Ruan, Yiyang Ma, Huan Yang, Huiguo He, Bei Liu, Jianlong Fu, Nicholas~Jing Yuan, Qin Jin, and Baining Guo.
\newblock Mm-diffusion: Learning multi-modal diffusion models for joint audio and video generation.
\newblock In \emph{Proceedings of the IEEE/CVF Conference on Computer Vision and Pattern Recognition}, pages 10219--10228, 2023.

\bibitem[Sanderson and Lovell(2009)]{sanderson2009vidtimit}
Conrad Sanderson and Brian~C Lovell.
\newblock Multi-region probabilistic histograms for robust and scalable identity inference.
\newblock In \emph{Advances in biometrics: Third international conference, ICB 2009, alghero, italy, june 2-5, 2009. Proceedings 3}, pages 199--208. Springer, 2009.

\bibitem[Siarohin et~al.(2019)Siarohin, Lathuili{\`e}re, Tulyakov, Ricci, and Sebe]{siarohin2019fomm}
Aliaksandr Siarohin, St{\'e}phane Lathuili{\`e}re, Sergey Tulyakov, Elisa Ricci, and Nicu Sebe.
\newblock First order motion model for image animation.
\newblock \emph{Advances in neural information processing systems}, 32, 2019.

\bibitem[Tong et~al.(2022)Tong, Song, Wang, and Wang]{tong2022videomae}
Zhan Tong, Yibing Song, Jue Wang, and Limin Wang.
\newblock Videomae: Masked autoencoders are data-efficient learners for self-supervised video pre-training.
\newblock \emph{Advances in neural information processing systems}, 35:\penalty0 10078--10093, 2022.

\bibitem[van~der Maaten and Hinton(2008)]{Maaten2008VisualizingDU}
Laurens van~der Maaten and Geoffrey~E. Hinton.
\newblock Visualizing data using t-sne.
\newblock \emph{Journal of Machine Learning Research}, 9:\penalty0 2579--2605, 2008.

\bibitem[Wang et~al.(2021{\natexlab{a}})Wang, Liu, Hu, Shi, and Mei]{wang2021facex}
Jun Wang, Yinglu Liu, Yibo Hu, Hailin Shi, and Tao Mei.
\newblock Facex-zoo: A pytorch toolbox for face recognition.
\newblock In \emph{Proceedings of the 29th ACM International Conference on Multimedia}, pages 3779--3782, 2021{\natexlab{a}}.

\bibitem[Wang et~al.(2021{\natexlab{b}})Wang, Mallya, and Liu]{wang2021one}
Ting-Chun Wang, Arun Mallya, and Ming-Yu Liu.
\newblock One-shot free-view neural talking-head synthesis for video conferencing.
\newblock In \emph{Proceedings of the IEEE/CVF conference on computer vision and pattern recognition}, pages 10039--10049, 2021{\natexlab{b}}.

\bibitem[Wang et~al.(2023)Wang, Bao, Zhou, Wang, and Li]{Wang_2023_CVPR}
Zhendong Wang, Jianmin Bao, Wengang Zhou, Weilun Wang, and Houqiang Li.
\newblock Altfreezing for more general video face forgery detection.
\newblock In \emph{Proceedings of the IEEE/CVF Conference on Computer Vision and Pattern Recognition (CVPR)}, pages 4129--4138, 2023.

\bibitem[Wodajo and Atnafu(2021)]{wodajo2021cvit}
Deressa Wodajo and Solomon Atnafu.
\newblock Deepfake video detection using convolutional vision transformer.
\newblock \emph{arXiv preprint arXiv:2102.11126}, 2021.

\bibitem[Wolf et~al.(2011)Wolf, Hassner, and Maoz]{wolf2011ytfaces}
Lior Wolf, Tal Hassner, and Itay Maoz.
\newblock Face recognition in unconstrained videos with matched background similarity.
\newblock In \emph{CVPR 2011}, pages 529--534. IEEE, 2011.

\bibitem[Wu et~al.(2023)Wu, Hui, and Zhou]{wu2023deepfake}
Haojie Wu, Pan Hui, and Pengyuan Zhou.
\newblock Deepfake in the metaverse: An outlook survey.
\newblock \emph{arXiv preprint arXiv:2306.07011}, 2023.

\bibitem[Yang et~al.(2023)Yang, Zhou, Chen, Guo, Ba, Xia, Cao, and Ren]{yang2023avoiddf}
Wenyuan Yang, Xiaoyu Zhou, Zhikai Chen, Bofei Guo, Zhongjie Ba, Zhihua Xia, Xiaochun Cao, and Kui Ren.
\newblock Avoid-df: Audio-visual joint learning for detecting deepfake.
\newblock \emph{IEEE Transactions on Information Forensics and Security}, 18:\penalty0 2015--2029, 2023.

\bibitem[Yang et~al.(2019)Yang, Li, and Lyu]{yang2019head}
Xin Yang, Yuezun Li, and Siwei Lyu.
\newblock Exposing deep fakes using inconsistent head poses.
\newblock In \emph{ICASSP 2019 - 2019 IEEE International Conference on Acoustics, Speech and Signal Processing (ICASSP)}, pages 8261--8265, 2019.

\bibitem[Yi et~al.(2020)Yi, Ye, Zhang, Bao, and Liu]{yi2020atfhp}
Ran Yi, Zipeng Ye, Juyong Zhang, Hujun Bao, and Yong-Jin Liu.
\newblock Audio-driven talking face video generation with learning-based personalized head pose.
\newblock \emph{arXiv preprint arXiv:2002.10137}, 2020.

\bibitem[Zakharov et~al.(2019)Zakharov, Shysheya, Burkov, and Lempitsky]{zakharov2019nth}
Egor Zakharov, Aliaksandra Shysheya, Egor Burkov, and Victor Lempitsky.
\newblock Few-shot adversarial learning of realistic neural talking head models.
\newblock In \emph{Proceedings of the IEEE/CVF international conference on computer vision}, pages 9459--9468, 2019.

\bibitem[Zhao et~al.(2022)Zhao, Yu, Ni, and Zhao]{zhao2022spatiotemporal}
Xiaohui Zhao, Yang Yu, Rongrong Ni, and Yao Zhao.
\newblock Exploring complementarity of global and local spatiotemporal information for fake face video detection.
\newblock In \emph{ICASSP 2022 - 2022 IEEE International Conference on Acoustics, Speech and Signal Processing (ICASSP)}, pages 2884--2888, 2022.

\bibitem[Zheng et~al.(2021)Zheng, Bao, Chen, Zeng, and Wen]{zheng2021ftcn}
Yinglin Zheng, Jianmin Bao, Dong Chen, Ming Zeng, and Fang Wen.
\newblock Exploring temporal coherence for more general video face forgery detection.
\newblock In \emph{Proceedings of the IEEE/CVF international conference on computer vision}, pages 15044--15054, 2021.

\bibitem[Zhou and Lim(2021)]{zhou2021avdfd}
Yipin Zhou and Ser-Nam Lim.
\newblock Joint audio-visual deepfake detection.
\newblock In \emph{Proceedings of the IEEE/CVF International Conference on Computer Vision}, pages 14800--14809, 2021.

\bibitem[Zhou et~al.(2018)Zhou, Xu, Landreth, Kalogerakis, Maji, and Singh]{zhou2018visemenet}
Yang Zhou, Zhan Xu, Chris Landreth, Evangelos Kalogerakis, Subhransu Maji, and Karan Singh.
\newblock Visemenet: Audio-driven animator-centric speech animation.
\newblock \emph{ACM Transactions on Graphics (TOG)}, 37\penalty0 (4):\penalty0 1--10, 2018.

\bibitem[Zhu et~al.(2022)Zhu, Wu, Zhu, Jiang, Tang, Zhang, Liu, and Loy]{zhu2022celebvhq}
Hao Zhu, Wayne Wu, Wentao Zhu, Liming Jiang, Siwei Tang, Li Zhang, Ziwei Liu, and Chen~Change Loy.
\newblock Celebv-hq: A large-scale video facial attributes dataset.
\newblock In \emph{European conference on computer vision}, pages 650--667. Springer, 2022.

\bibitem[Zhuang et~al.(2022)Zhuang, Chu, Tan, Liu, Yuan, Miao, Luo, and Yu]{zhuang2022uiavit}
Wanyi Zhuang, Qi Chu, Zhentao Tan, Qiankun Liu, Haojie Yuan, Changtao Miao, Zixiang Luo, and Nenghai Yu.
\newblock Uia-vit: Unsupervised inconsistency-aware method based on vision transformer for face forgery detection.
\newblock In \emph{Computer Vision – ECCV 2022: 17th European Conference, Tel Aviv, Israel, October 23–27, 2022, Proceedings, Part V}, page 391–407, Berlin, Heidelberg, 2022. Springer-Verlag.

\end{thebibliography}
}

\end{document}